\documentclass[10pt,twocolumn,letterpaper]{article}
\usepackage[pagenumbers]{cvpr}
\usepackage[T1]{fontenc}
\usepackage{colortbl,placeins}
\definecolor{cvprblue}{HTML}{386A91}
\usepackage[breaklinks,colorlinks,allcolors=cvprblue]{hyperref}
\title{Intrinsic Layers: Unifying Amodal Layering and Intrinsic Decomposition for Object Editing and Cross-Scene Reinsertion}
\newcommand{\AuthorNames}{Rui Chen, Yue Gong, Xiang Li, Dongliang Shao, Le Zhang, Jun Huang, Xiaochao Qu, Luoqi Liu, Junlin Hu, Ting Liu}
\newcommand{\AuthorDisplay}{%
  Rui Chen\textsuperscript{1,2,*}, Yue Gong\textsuperscript{1,2,*},
  Xiang Li\textsuperscript{2}, Dongliang Shao\textsuperscript{2}, Le Zhang\textsuperscript{2},\\
  Jun Huang\textsuperscript{2}, Xiaochao Qu\textsuperscript{2},
  Luoqi Liu\textsuperscript{2}, Junlin Hu\textsuperscript{1,\ensuremath{\dagger}},
  Ting Liu\textsuperscript{2,\ensuremath{\dagger}}}
\newcommand{\AffiliationDisplay}{%
  \textsuperscript{1}Beihang University\qquad\textsuperscript{2}MT Lab, Meitu Inc.}
\newcommand{\AuthorNotes}{%
  \begingroup
  \renewcommand{\thefootnote}{*}%
  \footnotetext[1]{Equal contribution. Work done during internships at MT Lab, Meitu Inc.}%
  \renewcommand{\thefootnote}{\ensuremath{\dagger}}%
  \footnotetext[2]{Corresponding authors: Junlin Hu
    (\href{mailto:hujunlin@buaa.edu.cn}{hujunlin@buaa.edu.cn}) and Ting Liu
    (\href{mailto:lt@meitu.com}{lt@meitu.com}).}%
  \endgroup}

\hypersetup{pdftitle={Intrinsic Layers: Unifying Amodal Layering and Intrinsic Decomposition for Object Editing and Cross-Scene Reinsertion},pdfauthor={\AuthorNames},bookmarksnumbered=true}
\AtBeginDvi{}
\begin{document}
\onecolumn
\thispagestyle{plain}
\vspace*{-28pt}
\begingroup
\centering
\begin{minipage}{0.95\linewidth}
\centering
\definecolor{titleink}{HTML}{22252A}
="[fonts/STIXTwoText-Semibold.ttf]:letterspace=0.8;extend=0.96" at 14.3pt
="[fonts/STIXTwoText-Medium.ttf]:letterspace=0.8;extend=0.96" at 14.3pt
="[fonts/STIXTwoText-Medium.ttf]" at 11.5pt
="[fonts/STIXTwoText-Medium.ttf]" at 10pt
\def\displaytitle#1: #2 for Object Editing #3\endtitle{%
  {\color{cvprblue}\hrule height 0.5pt}
  \vbox to 61pt{%
    \vfil
    {\color{titleink}\titleface\baselineskip=22pt
     \mbox{\raisebox{-3pt}[0pt][0pt]{\includegraphics[height=14.5pt]{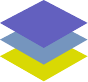}}\hspace{4.5pt}{\methodface #1:} #2}\\\mbox{for Object Editing #3}\par}
    \vfil}
  {\color{cvprblue}\hrule height 0.5pt}}
\expandafter\displaytitle\thetitle\endtitle
\ifx\AuthorNames\empty\else
  \vspace{9pt}{\fontsize{11.5}{15}\selectfont\authorface\AuthorDisplay\par}
  \vspace{5pt}{\sffamily\fontsize{10}{12}\selectfont
    \vphantom{\AffiliationDisplay}\smash{{\affiliationface\AffiliationDisplay}}\par}
\fi
\vspace{6mm}
\newsavebox{\abstractpanel}
\sbox{\abstractpanel}{\begin{minipage}{\linewidth}
  \vspace*{14pt}
  \centering
  \begin{minipage}{\dimexpr\linewidth-40pt\relax}
    \renewenvironment{abstract}{%
      {\centering\sffamily\bfseries\fontsize{12}{14}\selectfont
       \color{cvprblue}Abstract\par}%
      \vspace{7pt}\fontsize{11}{15.7}\selectfont
      \hyphenpenalty=400\doublehyphendemerits=10000\setlength{\emergencystretch}{1em}\noindent\ignorespaces
    }{\par}
    \begin{abstract}
Extracting editable object assets from a single image requires completing occluded regions, disentangling intrinsic properties, and removing object-induced background effects.
Existing layering methods retain environment-dependent appearance, while inverse rendering typically estimates visible-surface attributes without completing occluded object regions.
To address these limitations, we introduce IntrinsicPairs, a Blender-rendered dataset pairing object-present and object-absent scenes with amodal intrinsic annotations.
We propose \textsf{\textbf{IntrinsicLayers}}, a shared diffusion Transformer with material query conditioning that jointly recovers foreground, background, and complete intrinsic layers, including attributes of occluded object regions.
The companion IntrinsicInsert model synthesizes edited assets in target backgrounds and is frozen to provide insertion-response supervision during extraction training.
It handles transparent and emissive object interactions without explicit target-scene geometry or lighting, which conventional rendering of 2.5D assets typically requires.
Experiments demonstrate physically plausible amodal foreground completion, object removal, and intrinsic decomposition for material editing and cross-scene reinsertion.
\par\vspace{9pt}\noindent{\fontsize{9}{12}\selectfont \textsf{\textbf{Project page:}} \url{https://cr-1201.github.io/IntrinsicLayers/}\par}
\end{abstract}

  \end{minipage}\par
  \vspace*{15pt}
\end{minipage}}
\noindent\rlap{\raisebox{-\dp\abstractpanel}{%
  \includegraphics[width=\wd\abstractpanel,height=\dimexpr\ht\abstractpanel+\dp\abstractpanel\relax]{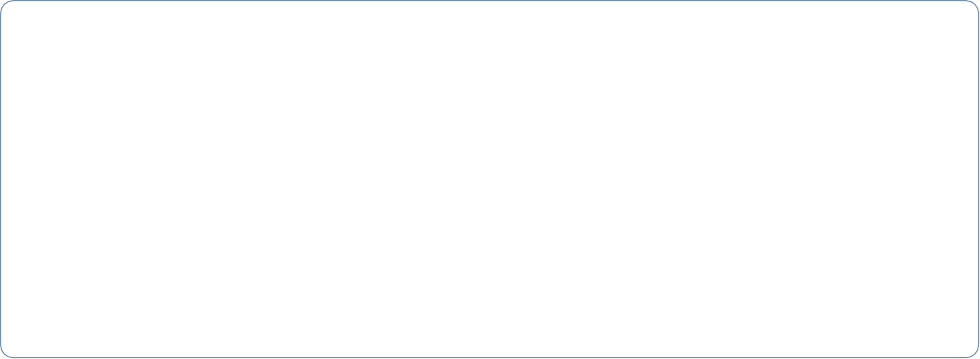}}}%
\usebox{\abstractpanel}\par
\vspace{12pt}
\makebox[\linewidth][c]{%
  \includegraphics[width=1.05263\linewidth,pagebox=cropbox]{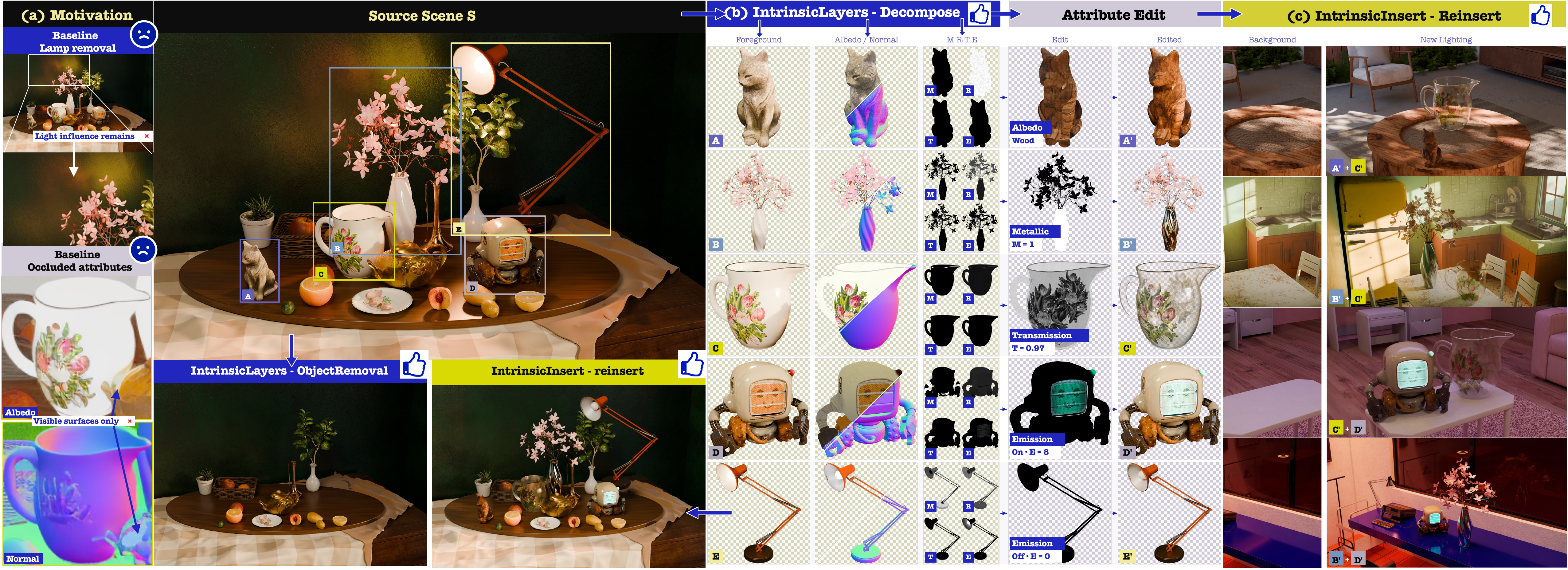}}\par
\captionsetup{font=small,labelfont={sf,bf},labelsep=period,skip=5pt}
\captionof{figure}{IntrinsicLayers extracts amodal object layers, intrinsic attributes, and a clean background from one image. These assets support material editing and cross-scene reinsertion, with shadows, reflections, transmission, and emission adapted to new RGB backgrounds.}
\label{fig:teaser}
\end{minipage}\par
\endgroup
\AuthorNotes
\twocolumn
\section{Introduction}
\label{sec:intro}

An object asset extracted from a single scene image should support independent editing and reuse in new backgrounds.
This requires a complete target-view projection with editable geometry and material attributes, together with a background free of the target and its visual effects.
After editing, the asset must adapt to a new background without retaining the original environment's lighting and reflections.

The main challenges are \textbf{object completeness} and \textbf{object--environment disentanglement}.
Occlusion hides parts of the target, so extracting a complete editable asset requires inferring missing shape and attributes while preserving visible details.
The observed object appearance contains reflections and lighting from the original environment, so directly reusing it in a new scene can produce unnatural and physically inconsistent results.
This coupling is particularly strong for transparent objects, whose appearance directly includes colors from the background.
A reusable background must also be free of the target's effects, including illumination from emissive objects.

Existing approaches address these requirements only in part.
Layer decomposition methods recover editable RGBA layers and can complete occluded objects~\cite{chen2026reflayer,yang2025layerdecomp}, but the extracted appearance remains coupled to the original environment.
They do not output intrinsic material maps, making subsequent material editing difficult.
Their recovered backgrounds can also retain reflections of the removed object and illumination from emissive targets (Figure~\ref{fig:fgbg_comparison}).
Image-space intrinsic decomposition estimates geometry and material attributes~\cite{li2025idarb}, but visible-surface estimates do not complete occluded target regions or recover an object-absent background.
Image-to-3D generation and NeRF- or Gaussian-based inverse rendering offer editable 3D representations~\cite{wang2024boosting3d,jin2023tensoir,liang2024gsir}; the former can lose fine geometric and texture details, while the latter require scene-specific optimization.
Physically rendering these assets into a new photograph still requires target-scene geometry and illumination to reproduce interactions such as cast shadows, beyond the background RGB image alone.
These approaches therefore do not directly provide joint recovery of a complete object, its intrinsic attributes, and a clean background for attribute-controlled editing and reinsertion.

To this end, we introduce \textbf{IntrinsicLayers}, which simultaneously performs (a) target-view foreground completion, (b) physically plausible object removal, including the object's effects on its surroundings, and (c) decomposition of the object's intrinsic geometry and material attributes.
Its companion model, \textbf{IntrinsicInsert}, takes the recovered or edited attribute maps and a target RGB background to synthesize physically plausible cross-scene insertions.
It handles background-dependent appearance for transparent objects and illumination from emissive objects without explicit lighting or background geometry and material maps.

To provide physically grounded supervision, we construct \textbf{IntrinsicPairs} by rendering each scene in Blender with and without the target from the same viewpoint, together with its amodal foreground and intrinsic attributes.
These counterfactual pairs supervise both object recovery and the removal of object-induced scene effects.
Our shared diffusion Transformer jointly denoises the background, foreground, and intrinsic layers, while learned material queries extract scene cues to condition metalness, roughness, and transmission prediction.
Training begins with supervised joint decomposition and subsequently adds insertion-response consistency, matching the frozen IntrinsicInsert model's single-step scene-velocity predictions under recovered and ground-truth attributes.
This constrains the extracted attributes through their combined effects on scene generation.
We construct \textbf{IntrinsicPairs-Test}, covering occluded, transparent, and emissive objects, to evaluate intrinsic recovery, layering, and object removal.

Our main contributions are as follows:
\begin{itemize}
    \item \textbf{Dataset:} IntrinsicPairs provides aligned counterfactual scene pairs, amodal foregrounds, and intrinsic attributes, supporting physically grounded supervision for layering, object removal, intrinsic decomposition, and insertion. The paired renders also supervise light transport effects from transparent and emissive targets.
    \item \textbf{Method:} IntrinsicLayers unifies amodal layering and intrinsic decomposition through joint layer prediction, material query conditioning, and insertion-response supervision from IntrinsicInsert, enabling attribute editing and cross-scene reinsertion.
    \item \textbf{Benchmark:} IntrinsicPairs-Test evaluates intrinsic recovery and foreground--background separation in scenes containing occlusion, transparency, and emission, including removal of the target's effects on its surroundings.
\end{itemize}

\begin{figure*}[t]
    \centering
    \includegraphics[width=\linewidth,pagebox=cropbox]{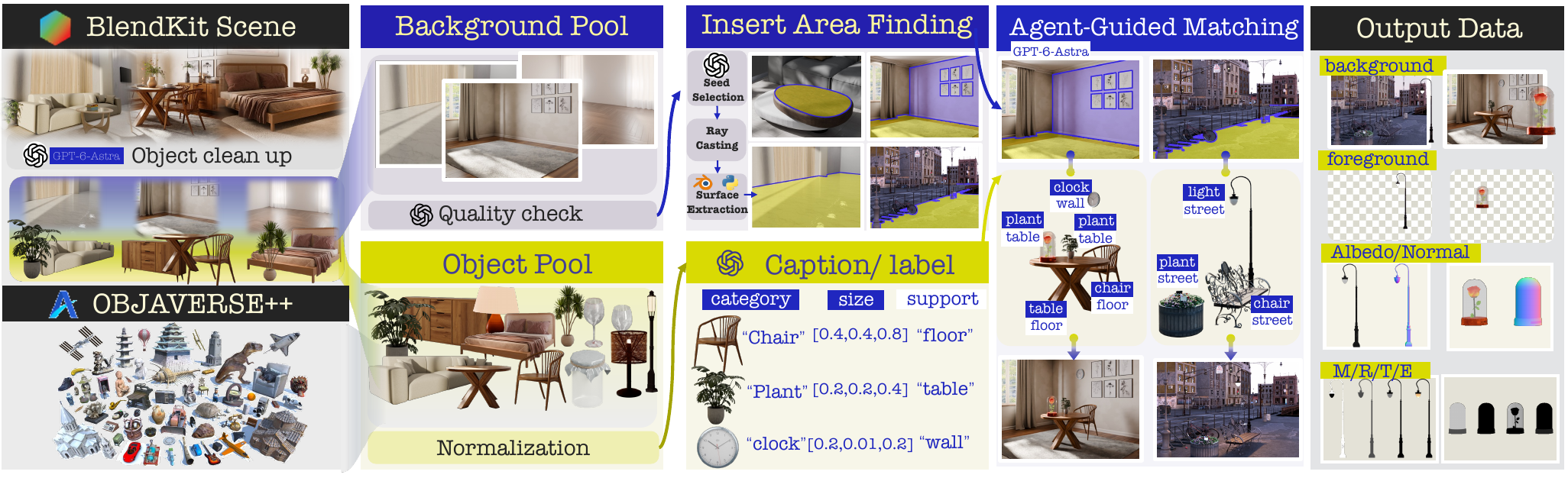}
    \caption{Data generation pipeline for IntrinsicPairs. We curate background scenes and normalize 3D assets. Ray casting identifies candidate support surfaces, and an agent matches assets to these surfaces by category, size, and support requirements. Rendering from a fixed camera produces paired scenes with and without the target, plus aligned amodal foreground RGBA and intrinsic maps.}
    \label{fig:data_pipeline}
\end{figure*}

\begin{figure}[t]
    \centering
    \includegraphics[width=\linewidth,pagebox=cropbox]{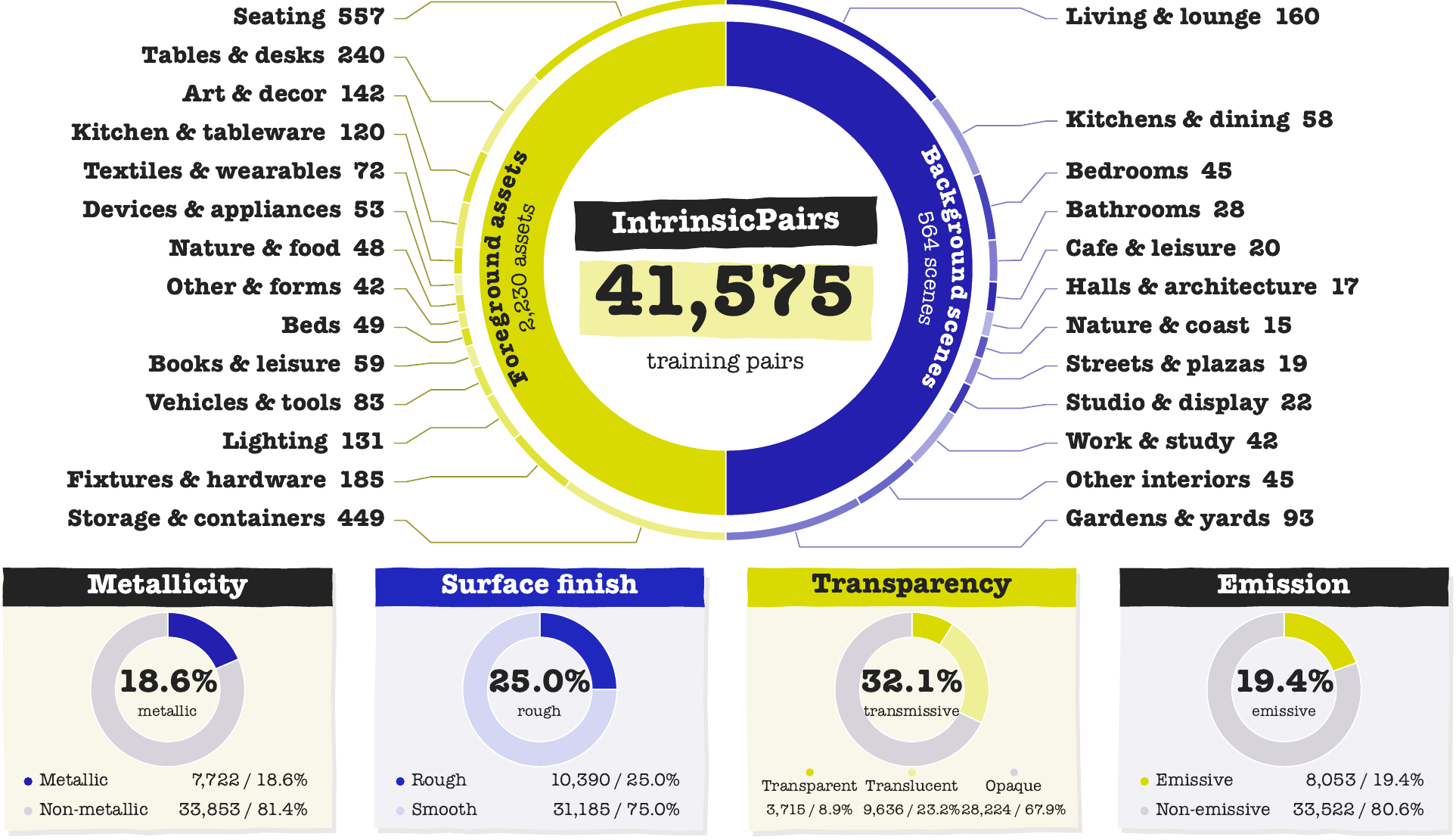}
    \caption{IntrinsicPairs statistics: category distributions of 2,230 foreground assets and 564 background scenes (top), and material frequencies over 41,575 training pairs (bottom).}
    \label{fig:dataset_overview}
\end{figure}

\begingroup
\section{Related Work}
\label{sec:related}

\paragraph{Image layer decomposition and amodal completion.}
Generative models recover or synthesize editable image layers, supporting foreground--background separation, transparency, and scene recomposition~\cite{yang2025layerdecomp,yin2025qwenimagelayered,wang2026reveallayer,yu2026omnialpha,zhang2024layerdiffuse}.
Amodal completion further recovers full target-view appearance from partial observations~\cite{ozguroglu2024pix2gestalt,chen2026reflayer,xu2024amodalcompletion,ao2025amodalappearance}.
Diffusion-based compositing and illumination editing also adapt reference objects to new scenes~\cite{chen2024anydoor,song2023objectstitch,zhang2025iclight}.
These image-level interfaces improve completeness and editability, but do not by themselves expose separate geometry and material controls: RGB content combines material properties with scene-dependent illumination and reflections, while compositing alpha does not specify physical transmission.
We pair amodal completion with aligned intrinsic attributes and a target-absent background, enabling attribute editing followed by cross-scene reinsertion.

\paragraph{Image-space 2.5D inverse rendering.}
Diffusion models estimate image-aligned intrinsic attributes from single or multiple views, with recent designs improving efficiency and detail preservation~\cite{kocsis2024intrinsicdiffusion,zeng2024rgbx,li2025idarb,hong2025supermat,wu2025stableintrinsic}.
Joint forward--inverse generation and rendering priors further constrain the decomposition~\cite{chen2025unirenderer,zheng2025dnfintrinsic}, while generative renderers extend attribute-conditioned synthesis to video~\cite{liang2025diffusionrenderer,zeng2026rgbxnext}.
Observed-view estimates do not by themselves form complete target-object assets: at occluded locations, scene predictions describe the visible occluder, so cropping cannot recover hidden target attributes.
Common attribute interfaces describe surface reflection without separate transmission and emission controls~\cite{zeng2024rgbx,li2025idarb,hong2025supermat}.
Although recent work includes transparency~\cite{zeng2026rgbxnext}, its image-aligned predictions do not jointly provide target-object amodal completion, independent emission attributes, and a target-absent background.
We organize these predictions into complete, editable object assets whose attributes support reinsertion with object--background interactions.

\paragraph{3D inverse rendering and material reconstruction.}
Optimization-based inverse rendering fits geometry, materials, or illumination to multi-view observations~\cite{zhang2021nerfactor,zhang2022invrender,munkberg2022nvdiffrec,jin2023tensoir,liang2024gsir,chen2024intrinsicanything}.
Specialized representations handle transparent objects and emissive sources~\cite{huang2025transparentgs,jeong2024esrnerf}, while single-image pipelines optimize predicted scenes for material editing~\cite{wang2026materialist}.
These editable representations require scene-specific test-time optimization.
With known geometry, diffusion priors complete unseen surface materials~\cite{wu2026matmart}.
Image-to-3D models generate textured or physically based assets~\cite{hunyuan2025hunyuan3d25,xiang2025trellis2}, while dedicated models complete geometry and appearance from occluded inputs~\cite{wu2025amodal3r}.
However, generated objects do not specify the target photograph's geometry, lighting, or occlusions; insertion requires additional scene estimation or image integration.
Our target-view extraction and insertion require no per-scene 3D fitting, object meshes, or background physical attributes.

\section{Method}
\label{sec:method}

IntrinsicLayers recovers an amodal foreground, intrinsic attributes, and a target-absent background from a scene image, target box, and text.
We construct IntrinsicPairs, a paired supervision dataset, and train a shared diffusion Transformer~\cite{peebles2023dit} on it through supervised decomposition followed by insertion-response consistency.

\begin{figure*}[t]
    \centering
    \includegraphics[width=\linewidth,pagebox=cropbox]{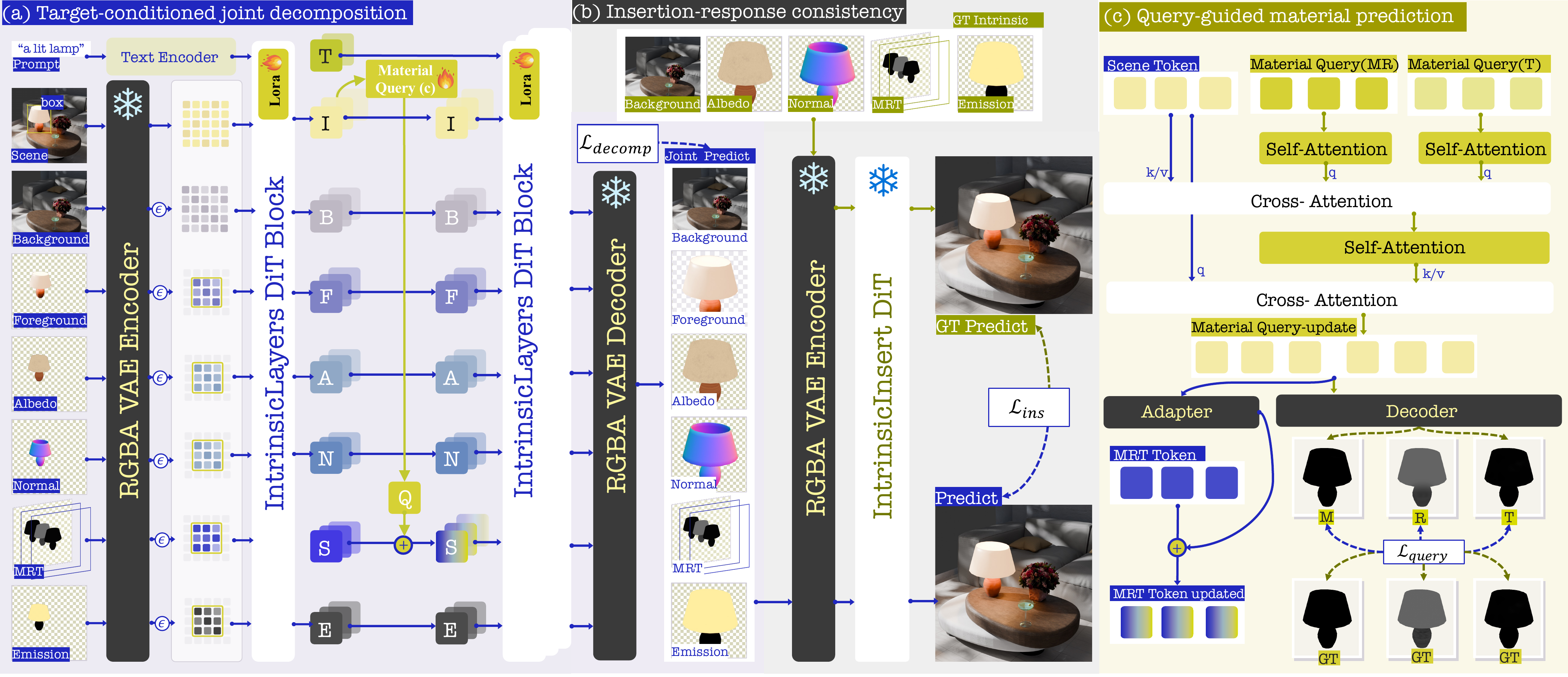}
    \caption{Overview of IntrinsicLayers. (a) Target-conditioned joint decomposition recovers background, foreground, and intrinsic layers. (b) Insertion-response consistency supervises predicted attributes through the frozen IntrinsicInsert model. (c) Material queries predict metalness, roughness, and transmission and condition the material stream.}
    \label{fig:method_overview}
\end{figure*}

\subsection{Dataset}
\label{sec:dataset}

To obtain physically consistent supervision for layering and intrinsic decomposition, we construct IntrinsicPairs through Blender rendering.
Figure~\ref{fig:data_pipeline} illustrates the data generation pipeline.
To diversify backgrounds while retaining plausible scene structure, an agent removes foreground objects from complete Blender scenes; the removed objects also expand our foreground asset pool.
The original scenes often contain complex object arrangements without clear foreground--background separation, making it difficult to control the layers used for training.

We therefore rebuild training scenes by reinserting objects into the cleaned backgrounds.
An agent identifies suitable insertion regions and uses object captions to match assets to compatible support surfaces based on category, size, and support requirements.
This produces diverse compositions with explicitly defined target objects.
From a fixed viewpoint, we render each composition with and without the target, together with its amodal foreground RGBA, geometric coverage, and intrinsic maps.
These aligned outputs jointly supervise foreground recovery, intrinsic decomposition, and removal of the target and its scene effects.

The 41,575 training pairs span 2,230 foreground assets and 564 background scenes (Figure~\ref{fig:dataset_overview}).
Transmissive and emissive targets occur in 32.1\% and 19.4\% of pairs, respectively, extending supervision beyond opaque objects.

\subsection{Target-Conditioned Joint Decomposition}
\label{sec:joint_decomposition}

Figure~\ref{fig:method_overview} summarizes the model and its training objectives.
The shared Transformer jointly denoises six aligned layers, retaining full-scene context while restricting object streams to the target region.

\paragraph{Aligned layer representation.}
Given a scene $I\in\mathbb R^{H\times W\times3}$, target box $b$, and text $c$, we predict layers indexed by $\mathcal K=\{B,F,A,N,S,E\}$: background, foreground, albedo, camera-space normals, packed materials, and RGB emission.
The foreground $F=(C_F,\alpha_F)$ uses straight RGBA, and $S=[M,R,T]$ packs metalness, roughness, and transmission.
A geometric coverage mask $G$ includes transparent surfaces in attribute supervision, independently of compositing alpha and transmission.

To reduce Transformer memory and focus object learning and supervision on the target, we retain only target-box tokens for the five object streams while keeping the full background stream.
Following ART~\cite{pu2025art}, we process object streams within the target region; in our implementation, cropping precedes frozen RGBA VAE encoding.
For each ground-truth layer $Y^{k,*}$, the selected latent tokens $z_0^k$ retain their global positional indices, preserving alignment across layers.
Scene and text tokens supply shared full-scene context, and role embeddings distinguish the six outputs (Figure~\ref{fig:method_overview}a).
Appendix~\ref*{app:token_selection} gives the cropping rule, token counts, and memory measurements.

\paragraph{Material query conditioning.}
We observe that direct DiT prediction often preserves clear material boundaries yet assigns incorrect metalness, roughness, or transmission values.
A dedicated material-query branch reads these parameters from the DiT's hidden scene features; its direct predictions provide more reliable values but less distinct boundaries.
We therefore use query-derived material features and values to condition the DiT's material flow-matching stream, combining explicit parameter estimation with detailed boundary generation.

Figure~\ref{fig:method_overview}c illustrates the query branch.
Because metalness and roughness often share spatial boundaries, we group their queries together and use a separate group for transmission.
The text- and box-conditioned queries first interact through self-attention within each group, then cross-attend to scene features $H_I$ from the DiT.
Joint self-attention subsequently exchanges information between the two groups.
Finally, scene tokens query the updated material tokens through cross-attention, producing spatial material features $U_{\mathrm{MR}}$ and $U_{\mathrm T}$.
A learnable decoder reads these features into a pixel-space material map $\widetilde S$, supervised against the ground-truth MRT map.
Writing the branch outputs as
\begin{equation}
    q=(U_{\mathrm{MR}},U_{\mathrm T},\widetilde S)
      =\mathcal Q_\psi\!\left(\operatorname{sg}(H_I),c,b\right),
    \label{eq:material_queries}
\end{equation}
we align and project the spatial features to the selected object tokens, concatenate them with $\widetilde S$, and inject the resulting condition into the MRT stream through adapters.
The final material map is generated through DiT denoising and frozen VAE decoding.
Here $\operatorname{sg}$ stops gradients; we also detach the query outputs before conditioning, so pixel-space material supervision trains the query branch while generative losses train the conditioning projections and adapters.
Implementation details appear in Appendix~\ref*{app:material_queries}.

\subsection{Stage I: Supervised Joint Decomposition}
\label{sec:decomposition_training}

All streams share a noise level $t$ with independent standard Gaussian noise $\epsilon^k$.
Following flow matching~\cite{lipman2023flowmatching}, the Transformer jointly predicts velocities from $z_t^k=(1-t)z_0^k+t\epsilon^k$, with targets $v^{k,*}=\epsilon^k-z_0^k$.
To express spatially weighted losses, define
\[
    \|V\|_{w,\delta}^2
    =\frac{\sum_j w_j\|V_j\|_2^2}
    {d_V\max\{\sum_j w_j,\delta\}},
\]
where $d_V$ is the channel dimension and $\delta$ is a denominator floor.
The supervised objective is
\begin{equation}
    \begin{aligned}
    \mathcal L_{\mathrm{flow}}
      &=\sum_{k\in\mathcal K}\lambda_k
        \|\widehat v_\theta^k-v^{k,*}\|_{w^k,0}^2,\\
    \mathcal L_{\mathrm{decomp}}
      &=\mathcal L_{\mathrm{flow}}
        +\sum_{k\in\mathcal K}\mu_k\mathcal L_{\mathrm{pix}}^k
        +\mu_q\mathcal L_{\mathrm{query}}.
    \end{aligned}
    \label{eq:decomp_loss}
\end{equation}
Background weights are uniform; object weights map coverage and boundary information to each token grid.
Each stream is normalized separately over its support.
The decoded losses use $\widehat Y^k=\mathcal D_k(z_t^k-t\widehat v_\theta^k)$, where $\mathcal D_k$ restores the spatial layout and applies the frozen decoder.

Color and alpha receive $L_1$ supervision for $B,F,A$, with background color evaluated globally, object color within $G$, and alpha over each layer's domain.
Normals use a cosine direction loss, a length regularizer, and an alpha loss; MRT supervision combines coverage-weighted $L_1$ with adjacent-pixel differences.
Emission supervision averages RGB $L_1$ over $G$ and nonzero emission when present.
The query-decoded map $\widetilde S$ uses coverage-weighted smooth $L_1$, upweights large metalness values, and adds binary cross-entropy on metalness.
Decoded terms are active only at their scheduled noise levels and intervals.

\subsection{Stage II: Insertion Response Consistency}
\label{sec:insertion_consistency}

Trained separately on the same IntrinsicPairs data, IntrinsicInsert generates complete scenes from a target background, placement box, text, and physical-attribute references.
Its objective combines target-weighted flow matching with intermittent decoded $L_1$ supervision.
IntrinsicInsert is frozen throughout Stage~II (Appendix~\ref*{app:insertion_model}).

Frozen IntrinsicInsert supervises object--environment interactions without the explicit scene geometry and lighting typically required by conventional 2.5D rendering (Figure~\ref{fig:method_overview}b).
Let $\widehat{\mathcal P}=\{\widehat A,\widehat N,\widehat S,\widehat E\}$ and $\mathcal P^*$ denote predicted and ground-truth attributes.
A shared encoder $\Gamma$ restores canvas coordinates, constructs masks and alpha from $G$, adapts formats and resolution, and encodes both sets into VAE references.
For IntrinsicInsert $v_\phi$, define the scene-velocity response
\begin{equation}
    u_s(\mathcal P)
    =v_\phi^{\mathrm{scene}}\!\left(\mathbf x_s,s;
        \Gamma(\mathcal P),c_I\right).
    \label{eq:insertion_velocities}
\end{equation}
Both calls share the noisy scene and auxiliary foreground state $\mathbf x_s$; $s$ is sampled independently of $t$.
Their conditions $c_I$ share the GT background, target location, $G$, and vision-language context computed once from detached predicted references.
We omit clean foreground RGB, so only the attribute references differ.
Figure~\ref{fig:method_overview}b illustrates these conditions; supervision matches single-step scene velocities.

With $\Delta u=u_s(\widehat{\mathcal P})-\operatorname{sg}(u_s(\mathcal P^*))$, consistency combines global and local response errors,
\begin{equation}
    \mathcal L_{\mathrm{ins}}
    =\frac{\|\Delta u\|_{\mathbf1,0}^2
        +\rho\|\Delta u\|_{w^I,1}^2}{1+\rho},
    \label{eq:insertion_loss}
\end{equation}
where $w^I$ projects dilated coverage onto scene tokens and $\rho$ weights the local term.
The denominator floor handles empty local support.
Stage II initializes from Stage I and retains its supervision,
\begin{equation}
    \mathcal L_{\mathrm{stage2}}=\mathcal L_{\mathrm{decomp}}
    +\lambda_{\mathrm{ins}}\mathbf1[t\in\mathcal T_{\mathrm{ins}}]
        \mathcal L_{\mathrm{ins}}.
    \label{eq:stage2_loss}
\end{equation}
Two single-step insertion evaluations reuse the decomposition forward pass.
Gradients reach predicted $A,N,S,E$ through frozen IntrinsicInsert, the reference encoder, and the decoder; $B,F$ retain decomposition supervision.
At inference, IntrinsicLayers alone jointly denoises all six streams and decodes the full background and five object layers at their original image coordinates.
Detailed input and reinsertion settings are provided in Appendix~\ref*{app:inference_inputs}.

\section{Experiments}
\label{sec:experiments}
\begin{figure*}[t]
\centering
\includegraphics[width=\linewidth,pagebox=cropbox]{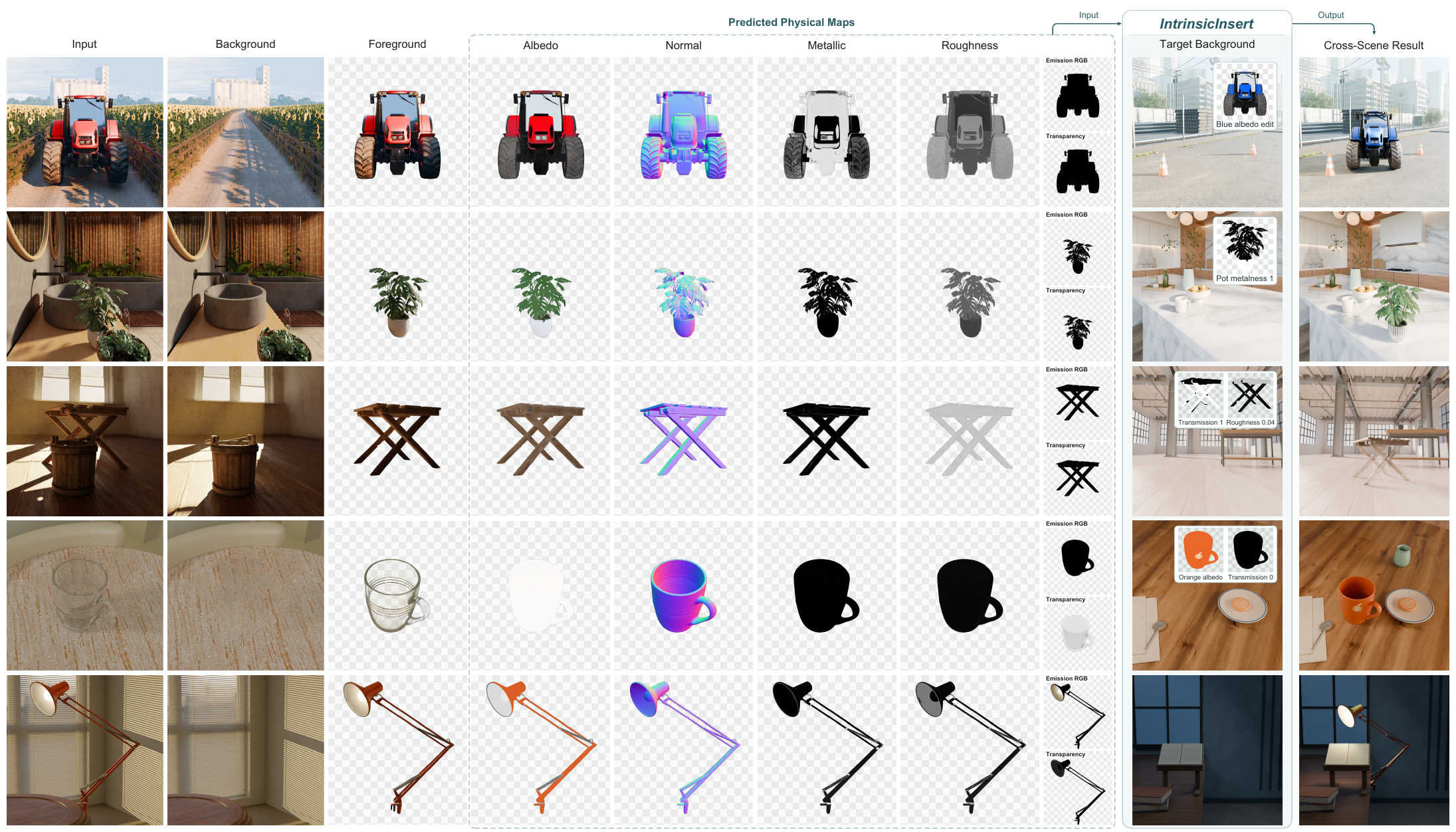}
\caption{Cross-scene reinsertion with the layers and physical attributes recovered by IntrinsicLayers. Insets on the target backgrounds show material edits, including a blue tractor, a metallic pot, transmissive stool legs, and an opaque orange mug. In the last row, the desk lamp illuminates its new surroundings.}
\label{fig:qualitative_cases}
\end{figure*}

\begin{table}[t]
\centering
\caption{Single-view inverse rendering on IntrinsicPairs-Test. Bold/underline mark best/second best throughout.}
\label{tab:intrinsic_results}
\fontsize{6}{7.2}\selectfont
\setlength{\tabcolsep}{0.65pt}
\renewcommand{\arraystretch}{1.18}
\resizebox{\columnwidth}{!}{%
\begin{tabular}{@{}l*{10}{c}@{}}
\toprule
\textbf{Method} & \multicolumn{3}{c}{\textbf{Albedo}} & \multicolumn{3}{c}{\textbf{Normal}} & \multicolumn{2}{c}{\textbf{Metalness}} & \multicolumn{2}{c}{\textbf{Roughness}} \\
\cmidrule(lr){2-4}\cmidrule(lr){5-7}\cmidrule(lr){8-9}\cmidrule(l){10-11}
& PSNR$^{\uparrow}$ & SSIM$^{\uparrow}$ & LPIPS$^{\downarrow}$ & Mean$^{\circ\downarrow}$ & 11.25$^{\circ\uparrow}$ & Cos.$^{\uparrow}$ & MSE$^{\downarrow}$ & MAE$^{\downarrow}$ & MSE$^{\downarrow}$ & MAE$^{\downarrow}$ \\
\midrule
RGB$\leftrightarrow$X & 13.45 & 0.637 & 0.346 & -- & -- & -- & 0.325 & 0.465 & 0.094 & 0.232 \\
IDArb-Single & 12.39 & 0.663 & 0.349 & 32.21 & 23.68 & 0.770 & 0.214 & 0.282 & 0.128 & 0.274 \\
IntrinsicAnything & 12.21 & 0.611 & 0.375 & -- & -- & -- & -- & -- & -- & -- \\
SuperMat & 14.71 & \underline{0.706} & \underline{0.321} & -- & -- & -- & \underline{0.097} & \underline{0.189} & \underline{0.063} & \underline{0.192} \\
Marigold V2 & 12.93 & 0.565 & 0.366 & 17.58 & \underline{59.14} & 0.896 & -- & -- & -- & -- \\
DNF-Intrinsic & 12.40 & 0.574 & 0.379 & -- & -- & -- & 0.117 & 0.201 & 0.146 & 0.312 \\
DiffusionRenderer & \underline{15.05} & 0.671 & 0.335 & 24.51 & 40.70 & 0.839 & 0.244 & 0.337 & 0.085 & 0.225 \\
RevealLayer\,+\,IDArb-Single & 14.33 & 0.534 & 0.344 & 27.84 & 34.30 & 0.809 & 0.197 & 0.290 & 0.146 & 0.291 \\
RevealLayer\,+\,Marigold V2 & 14.00 & 0.537 & 0.321 & \underline{15.88} & 56.74 & \underline{0.922} & -- & -- & -- & -- \\
\midrule
\textbf{IntrinsicLayers} & \textbf{21.56} & \textbf{0.774} & \textbf{0.145} & \textbf{15.07} & \textbf{60.20} & \textbf{0.929} & \textbf{0.053} & \textbf{0.091} & \textbf{0.038} & \textbf{0.139} \\
\bottomrule
\end{tabular}%
}
\end{table}

\subsection{Evaluation}
\label{sec:evaluation}

We evaluate intrinsic recovery, layer decomposition, and object removal on IntrinsicPairs-Test, complemented by Arb-Objaverse~\cite{li2025idarb,deitke2023objaverse}, TensoIR-Synthetic~\cite{jin2023tensoir}, Stanford-ORB~\cite{kuang2023stanfordorb}, and OBER-Test~\cite{zhao2026objectclear} for the corresponding tasks.
IntrinsicPairs-Test contains 300 samples, with both foreground assets and background scenes disjoint from those used for training.
All methods use a single input view, with each view processed independently.
All normal maps are represented and evaluated in camera coordinates.
Appendix~\ref*{app:experimental_details} details asset diversity, baseline configurations, and evaluation metrics.

\paragraph{Implementation details.}
We adapt Qwen-Image-2.1~\cite{qwen2026image21} with rank-64 LoRA~\cite{hu2022lora} using $1024\!\times\!1024$ scene images.
IntrinsicLayers is trained for 20k steps: 15k with decomposition supervision, followed by 5k adding insertion-response consistency through frozen IntrinsicInsert.
Ours denotes this model unless otherwise specified.
Inference uses 40-step Euler sampling; additional training settings are provided in Appendix~\ref*{app:training_settings}.

\subsection{Quantitative Results}
\label{sec:quantitative_results}

\paragraph{Joint intrinsic recovery.}
\label{sec:inverse_rendering_results}
\label{sec:joint_results}
Table~\ref{tab:intrinsic_results} evaluates intrinsic recovery on IntrinsicPairs-Test.
Since most baselines do not complete occluded regions, we also compare cascades that first complete the foreground with RevealLayer, then estimate its attributes with IDArb-Single or Marigold V2.
IntrinsicLayers achieves the best results across all ten metrics and outperforms both cascades on albedo and normals, reaching 21.56\,dB albedo PSNR versus 14.33\,dB for RevealLayer + IDArb-Single.

\paragraph{Inverse rendering on public benchmarks.}
\label{sec:public_inverse_rendering_results}
Table~\ref{tab:public_intrinsics} evaluates single-view inverse rendering on three public datasets: Arb-Objaverse, TensoIR-Synthetic, and Stanford-ORB.
IntrinsicLayers achieves competitive performance against dedicated inverse-rendering methods across the reported albedo, geometry, and material metrics.

\begin{table*}[t]
\centering
\caption{Single-view intrinsic estimation on three public benchmarks.}
\label{tab:public_intrinsics}
\scriptsize
\setlength{\tabcolsep}{1.25pt}
\renewcommand{\arraystretch}{1.12}
\resizebox{\textwidth}{!}{%
\begin{tabular}{@{}l*{17}{c}@{}}
\toprule
& \multicolumn{8}{c}{\textbf{Arb-Objaverse}} & \multicolumn{6}{c}{\textbf{TensoIR-Synthetic}} & \multicolumn{3}{c}{\textbf{Stanford-ORB}} \\
\cmidrule(lr){2-9}\cmidrule(lr){10-15}\cmidrule(l){16-18}
\textbf{Method} & \multicolumn{3}{c}{\textbf{Albedo}} & \multicolumn{3}{c}{\textbf{Normal}} & \textbf{Metalness} & \textbf{Roughness} & \multicolumn{3}{c}{\textbf{Albedo}} & \multicolumn{3}{c}{\textbf{Normal}} & \multicolumn{3}{c}{\textbf{Albedo}} \\
\cmidrule(lr){2-4}\cmidrule(lr){5-7}\cmidrule(lr){8-8}\cmidrule(lr){9-9}\cmidrule(lr){10-12}\cmidrule(lr){13-15}\cmidrule(l){16-18}
& PSNR$\uparrow$ & SSIM$\uparrow$ & LPIPS$\downarrow$ & Mean$^\circ\downarrow$ & $11.25^\circ\uparrow$ & Cos.$\uparrow$ & MSE$\downarrow$ & MSE$\downarrow$ & PSNR$\uparrow$ & SSIM$\uparrow$ & LPIPS$\downarrow$ & Mean$^\circ\downarrow$ & $11.25^\circ\uparrow$ & Cos.$\uparrow$ & PSNR$\uparrow$ & SSIM$\uparrow$ & LPIPS$\downarrow$ \\
\midrule
RGB$\leftrightarrow$X~\cite{zeng2024rgbx} & 30.43 & 0.952 & 0.071 & 73.71 & 9.16 & 0.214 & 0.313 & 0.119 & 15.94 & 0.577 & 0.109 & 34.32 & 20.78 & 0.749 & 39.21 & \underline{0.975} & 0.065 \\
IDArb-Single~\cite{li2025idarb} & 33.91 & 0.966 & 0.051 & 23.18 & 36.35 & 0.863 & 0.105 & \underline{0.075} & \underline{19.66} & 0.655 & \underline{0.062} & 23.09 & 29.24 & 0.877 & 39.78 & 0.972 & 0.064 \\
IntrinsicAnything~\cite{chen2024intrinsicanything} & 31.93 & 0.955 & 0.068 & -- & -- & -- & -- & -- & 18.46 & 0.624 & 0.097 & -- & -- & -- & 38.91 & 0.971 & 0.064 \\
SuperMat~\cite{hong2025supermat} & 32.26 & 0.965 & 0.063 & -- & -- & -- & 0.105 & 0.180 & 17.94 & 0.698 & 0.092 & -- & -- & -- & \textbf{40.22} & \textbf{0.976} & 0.065 \\
Marigold V2~\cite{pavlovic2026marigoldv2} & \textbf{34.23} & \underline{0.969} & \underline{0.047} & \textbf{16.27} & \textbf{47.83} & \textbf{0.930} & -- & -- & 18.21 & \underline{0.700} & 0.065 & \textbf{16.57} & \textbf{44.06} & \textbf{0.932} & 38.47 & 0.968 & \underline{0.062} \\
DNF-Intrinsic~\cite{zheng2025dnfintrinsic} & 26.76 & 0.903 & 0.086 & 44.04 & 7.78 & 0.659 & \underline{0.104} & 0.133 & 11.32 & 0.341 & 0.104 & 33.33 & 14.00 & 0.782 & 36.81 & 0.957 & 0.065 \\
DiffusionRenderer-SVD~\cite{liang2025diffusionrenderer} & 31.61 & 0.958 & 0.064 & 24.58 & 32.70 & 0.853 & 0.210 & 0.092 & \textbf{20.34} & 0.634 & 0.075 & 24.04 & 34.49 & 0.859 & 39.49 & 0.974 & 0.064 \\
\midrule
\textbf{IntrinsicLayers} & \underline{34.03} & \textbf{0.975} & \textbf{0.043} & \underline{21.43} & \underline{36.35} & \underline{0.885} & \textbf{0.104} & \textbf{0.075} & 19.16 & \textbf{0.745} & \textbf{0.057} & \underline{19.62} & \underline{41.64} & \underline{0.902} & \underline{40.20} & 0.970 & \textbf{0.059} \\
\bottomrule
\end{tabular}%
}
\end{table*}

\begin{table}[t]
\centering
\caption{Layer decomposition on IntrinsicPairs-Test. OmniAlpha$^*$ uses refmatting for foregrounds and object-clear for backgrounds.}
\label{tab:layer_results}
\scriptsize
\setlength{\tabcolsep}{2pt}
\renewcommand{\arraystretch}{1.12}
\begin{tabular*}{\columnwidth}{@{\extracolsep{\fill}}lcccccc@{}}
\toprule
\textbf{Method} & \multicolumn{3}{c}{\textbf{Background}} & \multicolumn{3}{c}{\textbf{Foreground}} \\
\cmidrule(lr){2-4}\cmidrule(l){5-7}
 & PSNR$\uparrow$ & SSIM$\uparrow$ & LPIPS$\downarrow$ & PSNR$\uparrow$ & RGB $L_1\downarrow$ & Soft IoU$\uparrow$ \\
\midrule
OmniAlpha$^*$ & \underline{31.95} & 0.890 & 0.157 & 27.06 & 0.084 & 0.740 \\
RevealLayer & 28.57 & \underline{0.898} & \underline{0.115} & \underline{31.20} & \underline{0.068} & \underline{0.860} \\
\midrule
\textbf{IntrinsicLayers} & \textbf{34.28} & \textbf{0.934} & \textbf{0.041} & \textbf{34.48} & \textbf{0.053} & \textbf{0.910} \\
\bottomrule
\end{tabular*}
\end{table}

\begin{table}[t]
\centering
\caption{Object removal on OBER-Test and IntrinsicPairs-Test. ObjectClear$^*$ is evaluated without AGF post-processing.}
\label{tab:removal_results}
\scriptsize
\setlength{\tabcolsep}{2pt}
\renewcommand{\arraystretch}{1.12}
\begin{tabular*}{\columnwidth}{@{\extracolsep{\fill}}lcccccc@{}}
\toprule
\textbf{Method} & \multicolumn{3}{c}{\textbf{OBER-Test}} & \multicolumn{3}{c}{\textbf{IntrinsicPairs-Test}} \\
\cmidrule(lr){2-4}\cmidrule(l){5-7}
 & PSNR$\uparrow$ & SSIM$\uparrow$ & LPIPS$\downarrow$ & PSNR$\uparrow$ & SSIM$\uparrow$ & LPIPS$\downarrow$ \\
\midrule
ObjectClear$^*$ & 27.63 & 0.768 & 0.083 & 27.67 & 0.818 & 0.196 \\
ObjectClear & \underline{32.04} & \underline{0.933} & \textbf{0.037} & \underline{30.55} & \underline{0.909} & 0.153 \\
RevealLayer & 30.42 & 0.920 & 0.067 & 28.57 & 0.898 & \underline{0.115} \\
\midrule
\textbf{IntrinsicLayers} & \textbf{33.29} & \textbf{0.953} & \underline{0.039} & \textbf{34.28} & \textbf{0.934} & \textbf{0.041} \\
\bottomrule
\end{tabular*}
\end{table}

\paragraph{Layer decomposition.}
\label{sec:layer_results}
Table~\ref{tab:layer_results} shows that IntrinsicLayers improves both foreground extraction and background recovery over OmniAlpha and RevealLayer.

\paragraph{Object removal.}
\label{sec:removal_results}
Table~\ref{tab:removal_results} evaluates the background independently against dedicated removal methods.
IntrinsicLayers leads all three metrics on IntrinsicPairs-Test, improving PSNR from 30.55\,dB for ObjectClear with AGF to 34.28\,dB.
On OBER-Test, it achieves the highest PSNR and SSIM, while ObjectClear retains a small LPIPS advantage, 0.037 versus 0.039.
Figure~\ref{fig:fgbg_comparison} further examines the removal of lighting and reflection effects outside the target region.

\begin{table*}[t]
\centering
\caption{Ablation study on IntrinsicPairs-Test. Baseline versus Stage~I+ evaluates material-query conditioning; Stage~I+ versus IntrinsicLayers evaluates insertion-response supervision. Rankings use unrounded scores.}
\label{tab:ablation}
\scriptsize
\setlength{\tabcolsep}{2pt}
\renewcommand{\arraystretch}{1.12}
\begin{tabular*}{\textwidth}{@{\extracolsep{\fill}}l*{13}{c}@{}}
\toprule
\textbf{Variant} & \multicolumn{3}{c}{\textbf{Albedo}} & \multicolumn{3}{c}{\textbf{Normal}} & \multicolumn{2}{c}{\textbf{Metalness}} & \multicolumn{2}{c}{\textbf{Roughness}} & \multicolumn{2}{c}{\textbf{Transmission}} & \textbf{Emission} \\
\cmidrule(lr){2-4}\cmidrule(lr){5-7}\cmidrule(lr){8-9}\cmidrule(lr){10-11}\cmidrule(lr){12-13}\cmidrule(l){14-14}
& PSNR$\uparrow$ & SSIM$\uparrow$ & LPIPS$\downarrow$ & Mean$^\circ\downarrow$ & $11.25^\circ\uparrow$ & Cos.$\uparrow$ & MSE$\downarrow$ & MAE$\downarrow$ & MSE$\downarrow$ & MAE$\downarrow$ & MSE$\downarrow$ & MAE$\downarrow$ & PSNR$\uparrow$ \\
\midrule
Baseline & 20.76 & \textbf{0.778} & \underline{0.144} & \underline{15.63} & \textbf{60.44} & \underline{0.919} & 0.056 & 0.110 & 0.049 & 0.155 & 0.016 & 0.048 & 44.86 \\
Stage~I & 20.98 & 0.770 & \textbf{0.142} & 16.59 & 54.75 & 0.915 & \underline{0.054} & 0.092 & \underline{0.040} & \underline{0.143} & 0.011 & 0.037 & 55.00 \\
Stage~I+ & \underline{21.02} & 0.771 & 0.150 & 15.90 & 57.66 & \underline{0.919} & 0.055 & \textbf{0.090} & 0.049 & 0.156 & \underline{0.009} & \underline{0.035} & \underline{57.70} \\
\midrule
\textbf{IntrinsicLayers} & \textbf{21.56} & \underline{0.774} & 0.145 & \textbf{15.07} & \underline{60.20} & \textbf{0.929} & \textbf{0.053} & \underline{0.091} & \textbf{0.038} & \textbf{0.139} & \textbf{0.009} & \textbf{0.034} & \textbf{59.50} \\
\bottomrule
\end{tabular*}
\end{table*}

\subsection{Qualitative Results}
\label{sec:qualitative_results}

\begin{figure}[t]
\centering
\includegraphics[width=\linewidth,pagebox=cropbox]{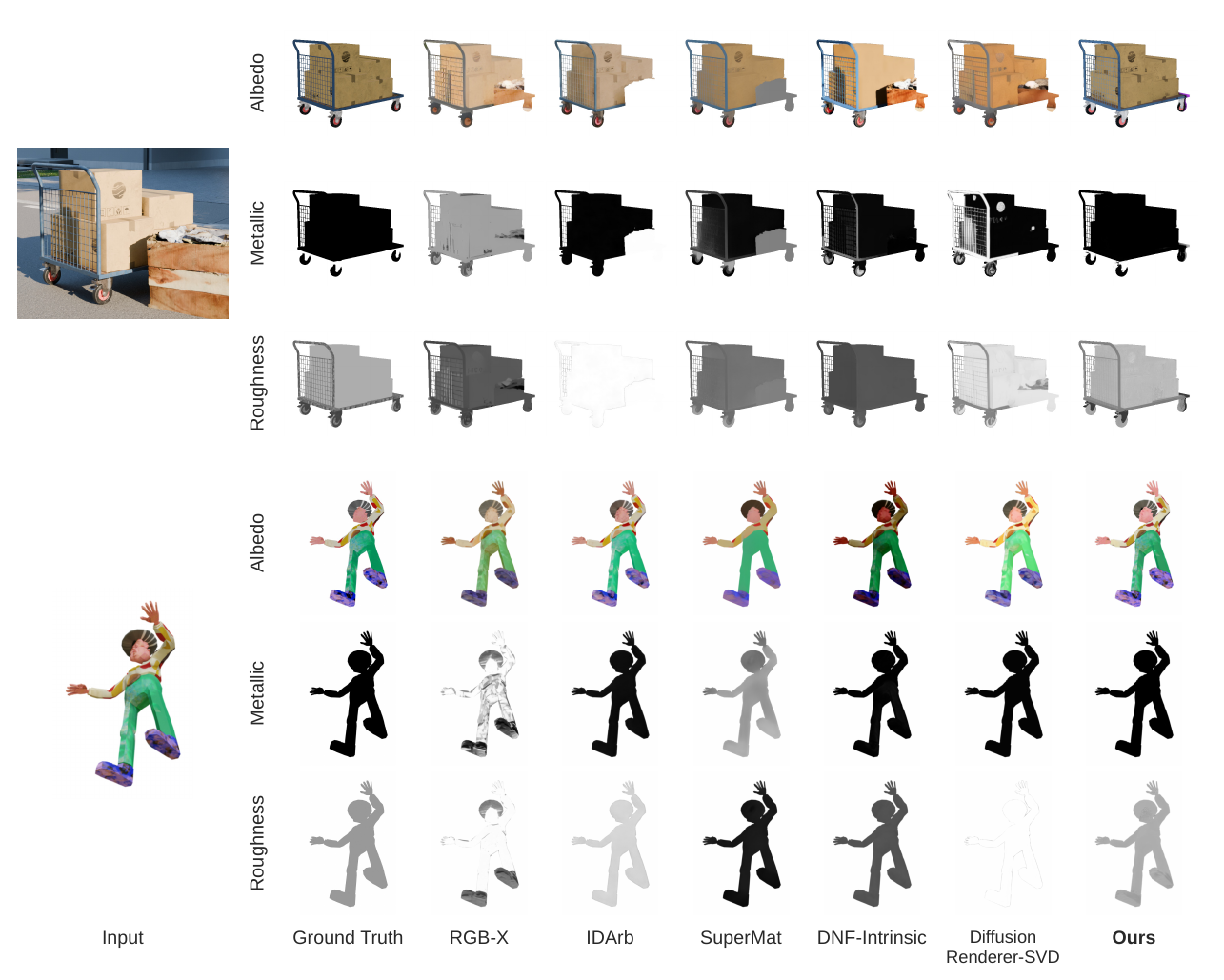}
\caption{Single-view albedo, metalness, and roughness estimation on IntrinsicPairs-Test (top) and Arb-Objaverse (bottom). Our prediction completes the cart's intrinsic maps behind the occluder.}
\label{fig:amr_comparison}
\end{figure}

\paragraph{Inverse rendering.}
In the first case of Figure~\ref{fig:amr_comparison}, the cart is partially occluded by a planter.
Among the compared methods, only IntrinsicLayers recovers the target's albedo, metalness, and roughness behind the occluder, demonstrating amodal attribute recovery beyond visible surfaces.
The Arb-Objaverse example shows that IntrinsicLayers preserves the figurine's clothing texture.
In Figure~\ref{fig:normal_comparison}, our suitcase normals more closely match ground truth in both fine surface structure and overall orientation.
Compared with DiffusionRenderer-SVD, we recover sharper vessel ridges and facial relief.
Further intrinsic comparisons appear in Figures~\ref{fig:app_amr_intrinsicpairs}--\ref{fig:app_normal_2}.

\begin{figure}[t]
\centering
\includegraphics[width=\linewidth,pagebox=cropbox]{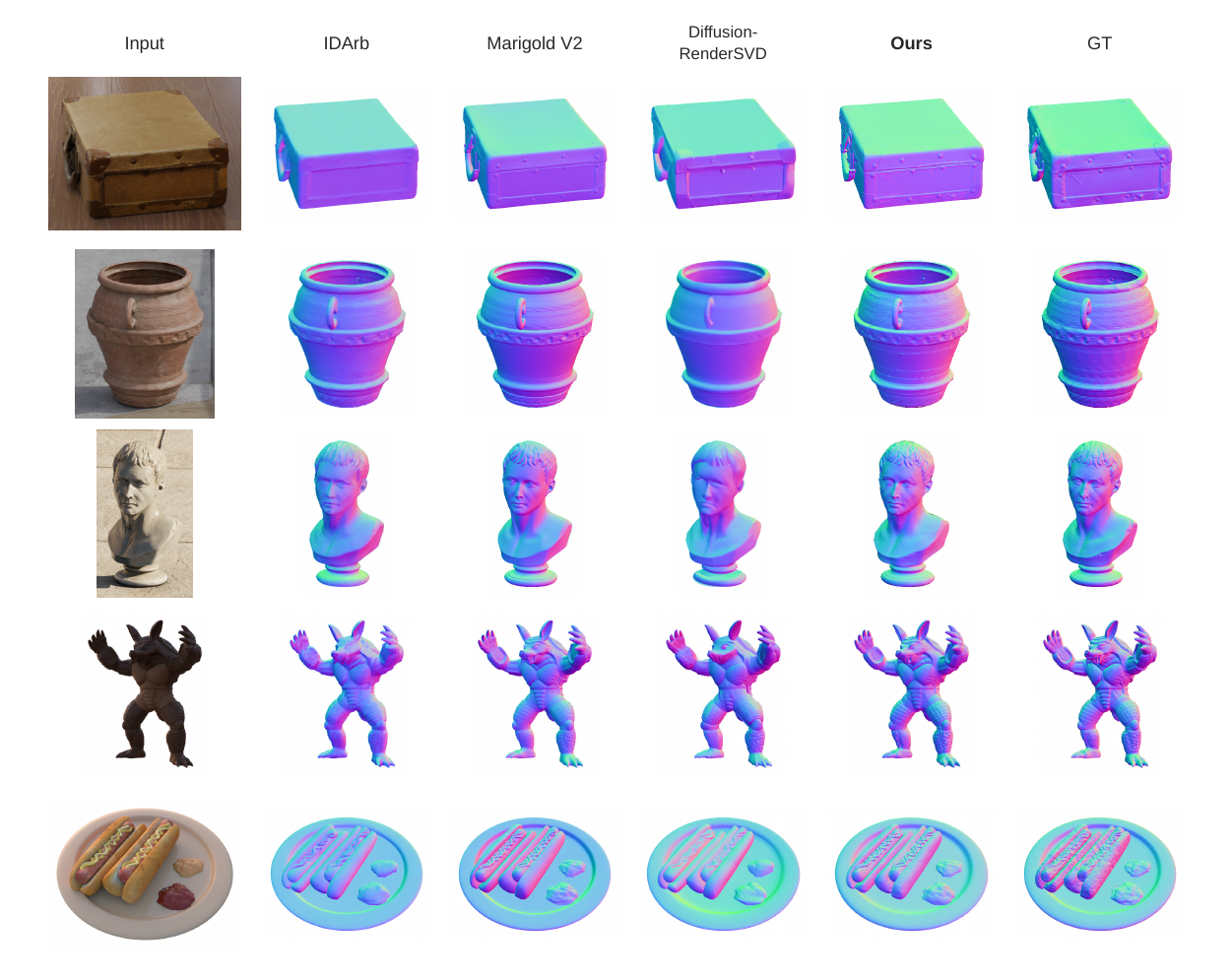}
\caption{Camera-space normal estimation on IntrinsicPairs-Test (top three rows) and public datasets (bottom two rows).}
\label{fig:normal_comparison}
\end{figure}

\paragraph{Object removal.}
The first two cases of Figure~\ref{fig:fgbg_comparison} test removal of emissive objects and their effects on the surrounding scene.
IntrinsicLayers removes the lamps together with the bright patches and illumination they cast on nearby surfaces, as well as their reflections in the metal sphere and glass panes.
This jointly restores the target region and the surrounding environment to a plausible object-absent appearance.
In comparison, ObjectClear and RevealLayer leave visible lighting and reflection residuals.

\begin{figure}[t]
\centering
\includegraphics[width=\linewidth,pagebox=cropbox]{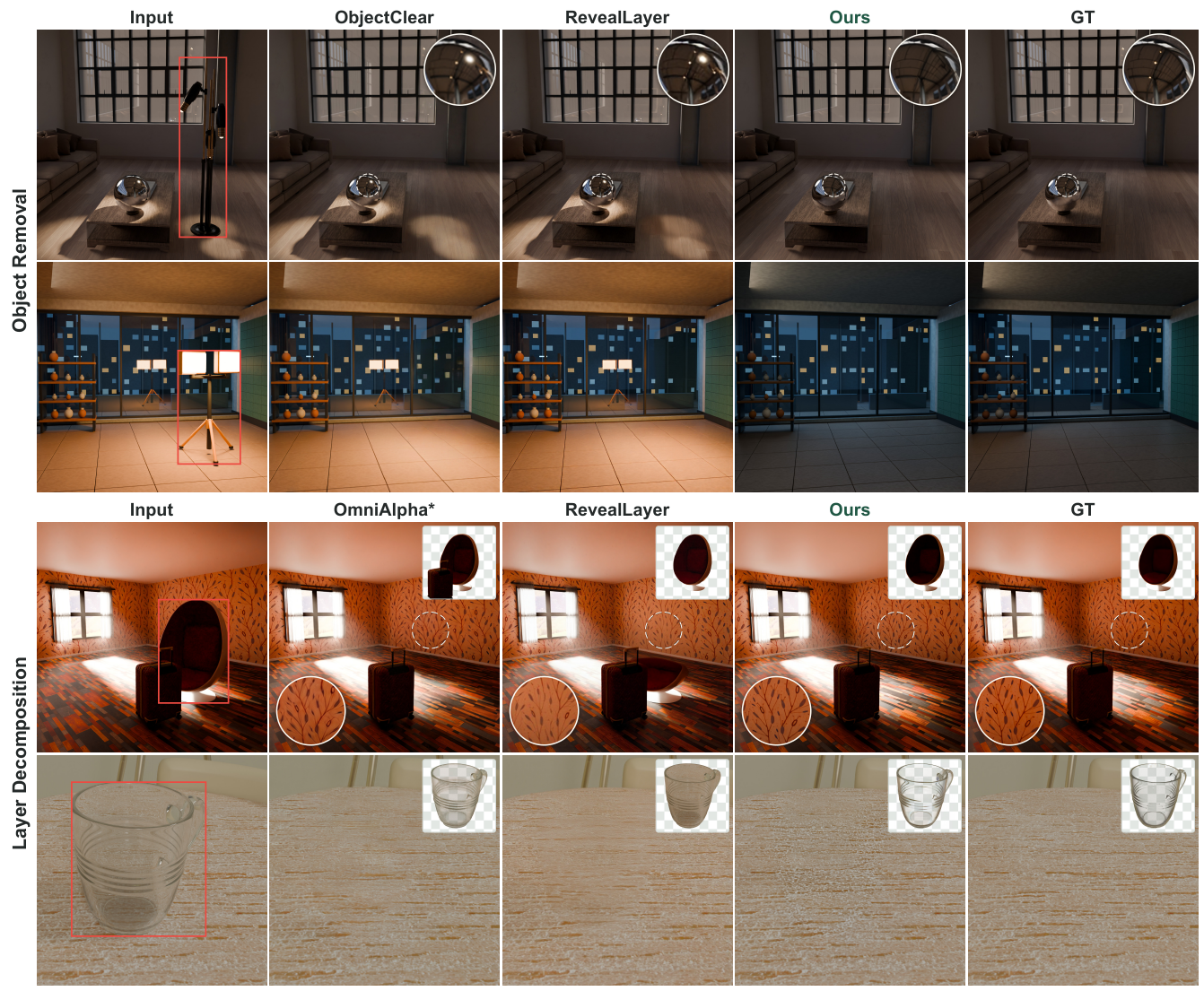}
\caption{Object removal (top two rows) and layer decomposition (bottom two rows). Our results remove target-induced shadows, reflections, and illumination, restore background texture, and preserve the glass foreground's transparency. OmniAlpha$^*$ uses the configuration in Table~\ref{tab:layer_results}.}
\label{fig:fgbg_comparison}
\end{figure}

\paragraph{Layer decomposition.}
The bottom two cases of Figure~\ref{fig:fgbg_comparison} demonstrate background texture recovery and clean foreground extraction.
In the chair scene, IntrinsicLayers restores the patterned wallpaper behind the target while separating the chair from the occluding suitcase.
For the transparent glass, our foreground preserves its outline and decorative bands without retaining the original tabletop texture in its interior, yielding a cleaner transparent layer for subsequent reuse.
These results complement the foreground and background improvements in Table~\ref{tab:layer_results}.
Additional joint layer and attribute outputs appear in Figures~\ref{fig:app_decomposition_1}--\ref{fig:app_decomposition_3}.

\paragraph{Object editing and reinsertion.}
Figure~\ref{fig:qualitative_cases} shows that IntrinsicInsert reinserts recovered objects into target RGB backgrounds after material editing, including changes in transmission.
In the last row, the lamp lights its new scene.
Additional results on insertion, lighting adaptation, and attribute control appear in Figures~\ref{fig:app_insertion_1}--\ref{fig:app_insertion_3} and~\ref{fig:app_lighting_sensitivity}--\ref{fig:app_emission_control}.
The user study is reported in Appendix~\ref{app:user_study}.

\subsection{Ablation Study}
\label{sec:ablation}

Table~\ref{tab:ablation} compares four models, all jointly predicting six layers with VAE-based MRT decoding.
Baseline trains for 20k steps without queries or insertion supervision.
Stage~I and Stage~I+ use queries with decomposition-only supervision for 15k and 20k steps.
IntrinsicLayers retains queries throughout 20k steps, adding insertion consistency only during the final 5k.
Appendix~\ref*{app:training_settings} provides details.

\paragraph{Material-query conditioning.}
Baseline and Stage~I+ both train for 20k steps without insertion supervision.
Stage~I+ reduces transmission MSE from 0.016 to 0.009 and metalness MAE from 0.110 to 0.090, while emission PSNR rises from 44.86 to 57.70\,dB.

\paragraph{Insertion-response consistency.}
Stage~I+ and IntrinsicLayers both use queries for 20k steps; only IntrinsicLayers adds insertion supervision during the final 5k.
Mean normal error falls from 15.90 to 15.07 degrees, roughness MSE from 0.049 to 0.038, and emission PSNR rises from 57.70 to 59.50\,dB.
Extending decomposition-only training from 15k to 20k improves emission but increases roughness MSE from 0.040 to 0.049.
These gains exceed those from longer decomposition training.

Emission may occupy only a small attribute-map region, such as a bulb, yet illuminate a much larger scene area.
Our decoded loss explicitly supervises nonzero-emission pixels but measures local attribute errors.
Through frozen IntrinsicInsert, missed emission can create scene-velocity discrepancies across many scene tokens, propagating gradients to the predicted emission.
This spatially distributed signal plausibly explains the additional emission gain over Stage~I+, complementing direct attribute supervision.

\FloatBarrier
\section{Conclusion}
\label{sec:conclusion}
We presented IntrinsicLayers for jointly recovering amodal object layers, intrinsic attributes, and a target-absent background from a single image.
IntrinsicPairs provides aligned supervision, while material queries guide prediction and a frozen IntrinsicInsert model supplies insertion-response supervision.
Experiments demonstrate accurate intrinsic recovery, layer decomposition, and object removal.
The recovered assets support material editing and cross-scene reinsertion with transmission and emission effects.

\section*{LLM Usage Statement}
We used OpenAI Codex to assist with drafting, organizing, and refining the manuscript text, including the abstract, introduction, related work, and descriptions of the method and experiments.
The research ideas were developed by the human authors, who designed and conducted all experiments and take full responsibility for the paper.

\endgroup
\clearpage
\appendix
\maketitlesupplementary

\setcounter{table}{0}
\setcounter{figure}{0}
\setcounter{equation}{0}
\renewcommand{\thetable}{S\arabic{table}}
\renewcommand{\thefigure}{S\arabic{figure}}
\renewcommand{\theequation}{S\arabic{equation}}
\renewcommand{\theHtable}{supp.\arabic{table}}
\renewcommand{\theHfigure}{supp.\arabic{figure}}
\renewcommand{\theHequation}{supp.\arabic{equation}}
\captionsetup[table]{position=top,labelsep=colon}
\captionsetup[figure]{labelsep=colon}

\section{Limitations}
\label{app:limitations}

Our 2.5D assets describe the object projection in the source view and do not support arbitrary-view rendering or 3D rotation.
Completing occluded regions and their intrinsic attributes relies on learned priors.
Because these hidden properties are not uniquely determined by a single image, plausible completions may differ from the true projected shape and material properties.

\section{Experimental Details}
\label{app:experimental_details}

\subsection{Datasets and Evaluation Protocols}
\label{app:evaluation_protocols}

IntrinsicPairs-Test comprises 300 paired object-present and object-absent images with amodal attributes, evaluating intrinsic recovery, layer decomposition, and object removal.
Both foreground assets and background scenes are disjoint from training.
OBER-Test~\cite{supp-zhao2026objectclear} additionally evaluates object removal.

The Arb-Objaverse subset~\cite{supp-li2025idarb,supp-deitke2023objaverse} contains 200 objects with four views each: 800 images processed independently by every method.
Scores use the original $512\!\times\!512$ ground truth and a fixed GT foreground region shared across methods.
TensoIR-Synthetic~\cite{supp-jin2023tensoir} evaluates albedo and surface normals.
Stanford-ORB~\cite{supp-kuang2023stanfordorb} evaluates albedo on 418 images aligned to $2048\!\times\!2048$ ground truth; PSNR follows the official linear-space, scale-aligned protocol.

\paragraph{Asset diversity.}
\label{app:asset_diversity}
The training pool contains 2,230 foreground assets in 14 categories and 564 background scenes in 12 categories (Figure~\ref*{fig:dataset_overview}).
Objects span furniture, lighting, household and natural objects, vehicles, and tools; scenes cover residential, work, leisure, and outdoor environments.
Of the 41,575 training pairs, 18.6\% contain metallic objects, 25.0\% rough surfaces, 32.1\% transmissive objects (8.9\% transparent and 23.2\% translucent), and 19.4\% emissive objects.
These statistics exclude the separate test asset pool.

The 300 test samples cover 164 distinct foreground target assets and 112 distinct background scene assets.
Foreground assets are deduplicated by target identifier, excluding non-target objects.
Backgrounds are deduplicated by source scene or traceable environment group; different viewpoints, target placements, and verified layout variants of the same environment count as one asset.

\subsection{Compared Methods}
\label{app:baseline_settings}

\paragraph{Inverse rendering.}
We compare with RGB$\leftrightarrow$X~\cite{supp-zeng2024rgbx}, IDArb~\cite{supp-li2025idarb}, IntrinsicAnything~\cite{supp-chen2024intrinsicanything}, SuperMat~\cite{supp-hong2025supermat}, Marigold V2~\cite{supp-pavlovic2026marigoldv2}, DNF-Intrinsic~\cite{supp-zheng2025dnfintrinsic}, and DiffusionRenderer-SVD~\cite{supp-liang2025diffusionrenderer}.
All use single-view inference.
Two cascades apply IDArb or Marigold V2 to foregrounds extracted by RevealLayer~\cite{supp-wang2026reveallayer}.
Table~\ref*{tab:intrinsic_results} abbreviates DiffusionRenderer-SVD as DiffusionRenderer.

\paragraph{Layer decomposition and removal.}
Layer decomposition compares RevealLayer and OmniAlpha~\cite{supp-yu2026omnialpha}.
Because OmniAlpha's automatic layering accepts neither a target box nor a mask, OmniAlpha$^*$ combines refmatting for foregrounds with object-clear for backgrounds (Table~\ref*{tab:layer_results}).
Removal compares RevealLayer and ObjectClear~\cite{supp-zhao2026objectclear}; ObjectClear includes AGF post-processing, while ObjectClear$^*$ omits it (Table~\ref*{tab:removal_results}).

\subsection{Evaluation Metrics}
\label{app:metrics}

\paragraph{Intrinsic attributes.}
Table~\ref{tab:app_metric_domains} specifies the domain of each metric on IntrinsicPairs-Test. Let $G_i(p)$ denote the ground-truth geometric coverage of pixel $p$ in image $i$. For a scalar error map $e_i$, define the within-image reduction
\begin{equation}
  \langle e_i\rangle_{G_i}
  =\frac{\sum_p G_i(p)e_i(p)}{\sum_p G_i(p)}.
  \label{eq:app_metric_reduction}
\end{equation}
For albedo, $e_i(p)=\|\widehat A_i(p)-A_i^*(p)\|_2^2/3$ in sRGB on $[0,1]$. We compute $-10\log_{10}\max(\langle e_i\rangle_{G_i},10^{-10})$ per image and average the 300 scores in dB. Thus a small object and a large object contribute equally to the dataset mean. For SSIM~\cite{supp-wang2004ssim} and LPIPS~\cite{supp-zhang2018lpips}, we multiply each albedo image by $G_i$ before computing the spatial score map and apply Eq.~\eqref{eq:app_metric_reduction} to that map.

\paragraph{Foreground and background.}
For foreground PSNR, each straight RGBA layer $(C,\alpha)$ yields a white-background composite $\alpha C+(1-\alpha)\mathbf{1}$. Foreground RGB $L_1$ uses straight colors, so transparent surfaces retain their contribution within $G>0$. We score opacity directly on the two full-resolution alpha maps using
\begin{equation}
  \mathrm{softIoU}_i =
  \frac{\sum_p \min(\widehat\alpha_i(p),\alpha_i^*(p))}
       {\max\!\left(10^{-12},\sum_p\max(\widehat\alpha_i(p),\alpha_i^*(p))\right)} .
  \label{eq:app_soft_iou}
\end{equation}

\paragraph{Reporting conventions.}
In the main-paper tables, PSNR, angular errors, and percentages use two decimal places; other metrics use three.
Bold and underlined scores rank first and second before rounding, so distinct ranks may display equal values.
Dashes indicate unreported results.

\subsection{Training and Ablation Configurations}
\label{app:training_settings}

\paragraph{Architecture and optimization.}
IntrinsicLayers adapts Qwen-Image-2.1~\cite{supp-qwen2026image21} with rank-64 LoRA~\cite{supp-hu2022lora} in its diffusion Transformer.
The pretrained Transformer, Qwen3-VL encoder~\cite{supp-bai2025qwen3vl}, and four-channel Alpha VAE are frozen; LoRA, role embeddings, material queries, conditioning projections, and adapters are trainable.
Scene images are $1024\!\times\!1024$.
Material conditioning follows Section~\ref*{sec:joint_decomposition}.
Table~\ref{tab:app_optimization} gives the AdamW~\cite{supp-loshchilov2019adamw} settings for the three training runs. In Stage~I the new conditioning projections and adapters follow a cosine decay after warmup, reaching the base rate at step 10,000 and retaining it through step 15,000. Stage~II resumes the Stage~I optimizer state on four NVIDIA L20X GPUs.

\paragraph{Target-region cropping.}
\label{app:token_selection}
Let $\mathcal E_{\mathrm{tok}}$ denote encoding and tokenization with the frozen RGBA VAE, and let $\mathcal R_\Omega$ crop the image region corresponding to token locations $\Omega$.
For ground-truth layer $Y^{k,*}$, we crop before VAE encoding to obtain the clean latent tokens
\begin{equation}
    z_0^k=\mathcal E_{\mathrm{tok}}(\mathcal R_{\Omega_k}(Y^{k,*})),
    \qquad
    \Omega_k=
    \begin{cases}
        \Omega_{\mathrm{full}}, & k=B,\\
        \Omega_b, & k\ne B.
    \end{cases}
    \label{eq:layer_streams}
\end{equation}
Here $\Omega_{\mathrm{full}}$ covers the full image, and $\Omega_b$ rounds the target box $b$ outward to the effective token stride $d$.
The selected tokens retain their global 3D RoPE~\cite{supp-su2024roformer} indices $(\ell_k,r_k+i,c_k+j)$, where $\ell_k$ identifies the layer and $(r_k,c_k)$ its grid origin.
For an image of size $H\times W$ and an aligned target region of size $h\times w$, selection reduces output tokens from $6HW/d^2$ to $(HW+5hw)/d^2$.
Cropping reduces both the object-layer pixels encoded by the VAE and the object tokens processed by the Transformer; the input scene and background output retain the full canvas.
With $d=16$, a $1024\times1024$ scene yields $64\times64$ spatial tokens. During training, we expand each side of the tight box by 10\% of its width or height, round outward to a 32-pixel grid, and clip to the canvas. The evaluation box uses the 5\% margin specified in Appendix~\ref{app:inference_inputs}.

Let $\bar r$ be the mean aligned box area divided by the full image area on the fixed-resolution canvas. Across the six output streams, the retained token fraction is $(1+5\bar r)/6$ and the saved fraction is $5(1-\bar r)/6$.

We measure the memory benefit with 36 paired training-step comparisons on one NVIDIA H20 (96\,GB), using batch size 1 and matched inputs, noise, and loss branches.
Peak allocated GPU memory is measured from online VAE encoding through the AdamW update.
After weighting by training-phase duration over the 20,000-step schedule, the mean peak decreases from 68.56 to 45.93\,GiB (33.0\%) with cropping.

\paragraph{Material-query implementation.}
\label{app:material_queries}
Our 32-block Transformer supplies 4096-dimensional scene features after block 8; prefix attention excludes noisy output tokens.
We project these features to 256 channels and select positions inside the target box. Six MR queries and three transmission queries exchange information through eight-head attention. The two spatial feature maps each supply 32 projected channels. Concatenating these with the three coarse material values gives a 67-dimensional condition per MRT token. The adapters before blocks 16, 24, and 32 each use a 128-dimensional bottleneck and a zero-initialized output projection.
A trainable operator $\mathcal C$ aligns the detached query outputs $q$ to $\Omega_b$ and concatenates projected spatial features with $\widetilde S$.
Adapters inject this condition only into the MRT stream:
\[
    H_{S,\mathrm{in}}^{(\ell)}\leftarrow H_{S,\mathrm{in}}^{(\ell)}
    +\mathcal A_\ell\!\left(\mathcal C(\operatorname{sg}(q))\right).
\]
The input scene features are also detached before entering $\mathcal Q_\psi$.
These two detachments restrict query-branch training to its pixel-space material loss while allowing generative losses to train $\mathcal C$ and the adapters.
The query decoder provides auxiliary material supervision and conditioning; it does not replace the frozen VAE decoder used for the final MRT output.

\paragraph{Decomposition supervision.}
We draw the shared noise level $t$ uniformly on $(0,1)$ and clip it to $[0.001,0.999]$. Each sample receives decoded MRT and query supervision. For adjacent-pixel MRT differences, the product of the two pixels' geometric coverage weights each horizontal or vertical pair. Let $\ell_M,\ell_R,\ell_T$ denote the three coverage-normalized smooth $L_1$ losses with transition point 0.05. Metalness regression uses $G(1+3M^*)$, and roughness and transmission use $G$. With $\ell_{\mathrm{BCE}}$ evaluated on metalness logits against continuous GT metalness, the query objective is
\begin{equation}
  \mathcal L_{\mathrm{query}}
  =(\ell_M+\ell_R+\ell_T+0.1\,\ell_{\mathrm{BCE}})/3 .
\end{equation}

Every fourth optimization step, when $t\in[0.1,0.7]$, the decoder supervises one of foreground, albedo, normals, and background in rotation. For the emission loss, emissive objects contribute two equally weighted RGB $L_1$ terms over $G$ and over the subset of covered pixels with nonzero GT emission. For non-emissive objects, the loss uses only the first term.

\paragraph{Training schedule and inference.}
Training uses 41,575 samples over 20k steps: 15k with decomposition and material supervision, then 5k adding same-scene insertion-response consistency.
The separately trained IntrinsicInsert remains frozen; only IntrinsicLayers is updated.
Both models use 40 flow-matching Euler steps at inference, with one conditional velocity prediction per step (guidance scale 1). Decomposition checkpoints use the same default seed of 42 for comparison.
Background decoding covers the full canvas; object layers are decoded locally and restored to their original image coordinates.

\paragraph{Insertion model.}
\label{app:insertion_model}
IntrinsicInsert uses the same IntrinsicPairs training data as IntrinsicLayers.
After rebuilding the optimizer, we continue training for 10k steps on four GPUs with global batch size 4.
The model generates the complete scene from a $1024\!\times\!1024$ target background, placement box, text, and $512\!\times\!512$ albedo, camera-space normal, MRT, and emission RGB references.
MRT is packed into one image and encoded with the frozen VAE; emission RGB is encoded separately.
The material stream uses a separate rank-32 LoRA. During independent inserter training, we drop the foreground RGB reference with probability 0.90 and jointly drop the object and material descriptions with probability 0.30. When the RGB reference is absent, an auxiliary stream predicts foreground RGB from noise.
The scene flow loss averages full-image and target-region errors with relative weights 1 and 4. Every fourth update, if the insertion noise level lies in $[0.2,0.6]$, we add decoded RGB $L_1$ with weight 0.1. Its scene term covers the object and a 32-pixel dilation around it.
All IntrinsicInsert parameters are then frozen for same-scene consistency training of IntrinsicLayers.

\paragraph{Stage~II response supervision.}
In Eq.~\eqref{eq:stage2_loss}, $\mathcal T_{\mathrm{ins}}=[0.1,0.5]$ gates the loss on the decomposition noise level $t$. For each eligible sample, we draw an independent insertion noise level $s\in[0.3,0.7]$ from the inserter's 1,000-point shifted schedule and reuse it for both attribute sets. To form $w^I$ in Eq.~\eqref{eq:insertion_loss}, we dilate GT coverage by 32 pixels and average it onto the $64\times64$ scene-token grid. We use $\rho=4$ and $\lambda_{\mathrm{ins}}=0.0788$. The latter was calibrated on 14 examples spanning different material properties by comparing gradient norms of the decomposition and consistency losses on their shared trainable parameters.

\paragraph{Ablation configurations.}
All variants jointly predict six layers with VAE-based MRT prediction.
Baseline trains for 20k steps without queries or insertion supervision.
Stage~I and Stage~I+ use queries without $\mathcal L_{\mathrm{ins}}$ for 15k and 20k steps.
IntrinsicLayers keeps queries throughout: 15k decomposition-only steps, then 5k adding $\mathcal L_{\mathrm{ins}}$ through frozen IntrinsicInsert.

\subsection{Inference Inputs and Reinsertion}
\label{app:inference_inputs}

\paragraph{Target boxes and baseline inputs.}
Decomposition takes a scene, target description, and a supplied box enclosing the complete target, including occluded regions.
For annotated decomposition evaluation, we compute the tight box from nonzero geometric coverage, expand each side by 5\% of its width or height, round outward to the 16-pixel token grid, and clamp to the image.
RevealLayer uses the same tight amodal box with its official grid alignment.
ObjectClear and OmniAlpha's ObjectClear task receive the full geometric target mask binarized at 0.5.
OmniAlpha's refmatting task uses the scene and target description.

\paragraph{Coverage for reinsertion.}
IntrinsicInsert takes a chosen target background, placement box, text, and the recovered attribute references.
These RGBA references carry predicted geometric support in their alpha channels.
When repositioning the references together, predicted albedo alpha supplies the shared support for alpha-aware resampling.
References retained at their original coordinates preserve their respective predicted alpha channels.
The ground-truth coverage $G$ in Section~\ref*{sec:insertion_consistency} supplies the shared conditions for the paired Stage~II training evaluations.

\section{Same-Scene Fidelity of Frozen IntrinsicInsert}
We evaluate frozen IntrinsicInsert on 500 cases from the IntrinsicPairs training set. For each case, we reinsert the object into its original background using ground-truth albedo, normals, MRT, and emission. Material text is enabled and the foreground RGB reference is disabled.
For comparison, we encode and decode the ground-truth object-present image with the RGBA VAE. We score both outputs against the original object-present image over the full frame (Table~\ref{tab:app_insertion_fidelity}).

\begin{table}[!htbp]
    \centering
    \small
    \caption{\textbf{Same-scene insertion fidelity.} Full-image scores on 500 IntrinsicPairs training cases.}
    \label{tab:app_insertion_fidelity}
    \setlength{\tabcolsep}{4pt}
    \renewcommand{\arraystretch}{1.10}
    \begin{tabular*}{\columnwidth}{@{\extracolsep{\fill}}lrrr@{}}
        \toprule
        \textbf{Output} & PSNR (dB)$\uparrow$ & SSIM$\uparrow$ & LPIPS$\downarrow$ \\
        \midrule
        IntrinsicInsert (GT attributes) & 31.68 & 0.9510 & 0.0398 \\
        RGBA VAE reconstruction & 42.63 & 0.9762 & 0.0107 \\
        \bottomrule
    \end{tabular*}
\end{table}

\section{Training and Evaluation Settings}
Table~\ref{tab:app_optimization} compares the two decomposition stages with independent inserter training. All runs use AdamW with $\beta=(0.9,0.999)$, $\epsilon=10^{-8}$, and a gradient norm cap of 1. Each of four GPUs processes one scene per update. Forward passes use BF16, and optimizer states use FP32.

Both IntrinsicLayers stages use weight decay 0.01. Stage~II retains the Stage~I losses and adds response consistency with $s\in[0.3,0.7]$ and local weight $\rho=4$. In IntrinsicInsert, type and role embeddings share a learning rate. Table~\ref{tab:app_metric_domains} specifies the representation and scoring region used for each output.

\begin{table*}[!t]
    \centering
    \caption{\textbf{Training hyperparameters.} IntrinsicLayers resumes the Stage~I optimizer for Stage~II. IntrinsicInsert is trained independently and frozen during Stage~II.}
    \label{tab:app_optimization}
    \small
    \renewcommand{\arraystretch}{1.15}
    \begin{tabular*}{\textwidth}{@{\extracolsep{\fill}}lccc@{}}
        \toprule
         & \multicolumn{2}{c}{\textbf{IntrinsicLayers}} & \textbf{IntrinsicInsert} \\
        \cmidrule(lr){2-3}\cmidrule(l){4-4}
        \textbf{Setting} & \textbf{Stage I} & \textbf{Stage II} & \textbf{Independent training} \\
        \midrule
        Optimizer updates & 15,000 & 5,000 & 10,000 \\
        Image-stream LoRA LR & $6\times10^{-6}$ & $3\times10^{-6}$ & $5\times10^{-6}$ \\
        Role embedding LR & $6\times10^{-6}$ & $3\times10^{-6}$ & $1\times10^{-5}$ \\
        Material query LR & $6\times10^{-6}$ & $3\times10^{-6}$ & --- \\
        Projection and adapter LR (peak to final) & $3\times10^{-5}\to6\times10^{-6}$ & $3\times10^{-6}$ & --- \\
        Material-stream LoRA LR & --- & --- & $1\times10^{-5}$ \\
        \addlinespace[2pt]
        LR schedule & 100-step warmup & Constant & Constant \\
        \midrule
        Auxiliary supervision & Decoded layer losses & Response consistency & Decoded RGB $L_1$ \\
        Activation & Layer-specific (Sec.~\ref{app:training_settings}) & $t\in[0.1,0.5]$ & Every 4 steps, $s\in[0.2,0.6]$ \\
        Auxiliary loss weight & Layer-specific & $\lambda_{\mathrm{ins}}=0.0788$ & 0.1 \\
        \bottomrule
    \end{tabular*}
\end{table*}

\begin{table*}[!t]
    \centering
    \caption{\textbf{Evaluation protocol on IntrinsicPairs-Test.} Each metric is computed per image and averaged over 300 cases. $G$ denotes ground-truth geometric coverage and $P_{11.25}$ the fraction of normals within $11.25^\circ$.}
    \label{tab:app_metric_domains}
    \small
    \renewcommand{\arraystretch}{1.15}
    \begin{tabular*}{\textwidth}{@{\extracolsep{\fill}}llll@{}}
        \toprule
        \textbf{Output} & \textbf{Metric} & \textbf{Representation} & \textbf{Evaluation region} \\
        \midrule
        Albedo & PSNR$\uparrow$ & sRGB & Weighted by $G$ \\
               & SSIM$\uparrow$, LPIPS$\downarrow$ & $G$-masked sRGB & Score maps weighted by $G$ \\
        Normals & Mean angle$\downarrow$, $P_{11.25}\uparrow$, cosine$\uparrow$ & Unit camera-space normals & Valid GT normals, weighted by $G$ \\
        MRT & MSE$\downarrow$, MAE$\downarrow$ & Per-channel material values & Weighted by $G$ \\
        Emission & PSNR$\uparrow$ & Log-encoded RGB & Weighted by $G$ \\
        \midrule
        Foreground & PSNR$\uparrow$ & RGB composited on white & Full image \\
                   & RGB $L_1\downarrow$ & Straight RGB & Pixels with $G>0$ \\
                   & Soft IoU$\uparrow$ & Continuous alpha & Full image \\
        Background & PSNR$\uparrow$, SSIM$\uparrow$, LPIPS$\downarrow$ & Object-absent RGB & Full image \\
        \bottomrule
    \end{tabular*}
\end{table*}

\section{Cross-Scene Transfer on Rendered Pairs}
\label{app:cross_scene_rendered}
\paragraph{Task and reference.}
Given an object in a source scene and a placement box in a target scene, we transfer the object at a fixed view using image-plane translation and uniform scaling. Each of the 100 pairs has a Blender render of the target scene with the object in place. The set comprises 62 standard transfers, 18 transfers of emissive objects, 12 transfers across lighting conditions, and eight transfers involving occlusion. The 100 evaluation pairs do not overlap with the training set. Each pair contributes equally to the mean for every method.

\paragraph{Method inputs.}
IntrinsicLayers recovers albedo, normals, MRT, and emission from the source. IntrinsicInsert then uses these attributes to generate the target scene. To measure the effect of attribute recovery, we also run the same inserter with GT attributes and score its output against the same Blender render. The five image editors receive identical source and target images marked with the selection and placement boxes. A shared prompt specifies fixed pose, object identity, and adaptation to target lighting. MiniMax applies additional prompt enhancement internally.

\paragraph{Full-image and local scores.}
We compare display-space RGB at $1024\times1024$. Kling's native $2048\times2048$ output is downsampled with Lanczos filtering. All comparisons use the target render's coordinate system without registration or color fitting. For the local scores, we expand each side of the GT geometry box by 10\% of its longer side, with at least 16 pixels of padding, enforce a minimum crop size of $64\times64$, and clip to the canvas. We score this crop at native scale, including its surrounding context. Both regions use RGB PSNR, SSIM with an $11\times11$ Gaussian window ($\sigma=1.5$), and AlexNet LPIPS v0.1.

\begin{table*}[!t]
    \centering
    \caption{\textbf{Cross-scene insertion on 100 rendered pairs.} All seven outputs are scored against the same Blender references. Bold marks the best result among methods operating on source RGB. The final row gives the GT-attribute control.}
    \label{tab:app_cross_scene}
    \small
    \renewcommand{\arraystretch}{1.13}
    \begin{tabular*}{\textwidth}{@{\extracolsep{\fill}}lrrrrrr@{}}
        \toprule
        & \multicolumn{3}{c}{\textbf{Full image}} & \multicolumn{3}{c}{\textbf{Target crop}} \\
        \cmidrule(lr){2-4}\cmidrule(l){5-7}
        \textbf{Method} & PSNR (dB)$\uparrow$ & SSIM$\uparrow$ & LPIPS$\downarrow$ & PSNR (dB)$\uparrow$ & SSIM$\uparrow$ & LPIPS$\downarrow$ \\
        \midrule
        Nano Banana Pro & 25.254 & 0.8780 & 0.1107 & 19.648 & 0.7483 & 0.2312 \\
        Seedream 5.5 & 21.167 & 0.7538 & 0.1427 & 15.942 & 0.5791 & 0.3481 \\
        GPT Image 2.6 Sunburst & 21.890 & 0.8298 & 0.1371 & 15.396 & 0.5989 & 0.3243 \\
        MiniMax H3 Image & 22.502 & 0.8092 & 0.1483 & 17.702 & 0.6826 & 0.2449 \\
        Kling Image 4.0 & 21.320 & 0.8068 & 0.1623 & 15.078 & 0.6003 & 0.4027 \\
        \midrule
        Ours (predicted attributes) & \textbf{29.249} & \textbf{0.9385} & \textbf{0.0482} & \textbf{22.715} & \textbf{0.8344} & \textbf{0.1304} \\
        Ours (GT attributes) & 29.500 & 0.9398 & 0.0455 & 23.092 & 0.8399 & 0.1175 \\
        \bottomrule
    \end{tabular*}
\end{table*}

\paragraph{Results.}
With predicted attributes, our pipeline exceeds the strongest image editor, Nano Banana Pro, by 3.99\,dB in full-image PSNR and 3.07\,dB within the target crop (Table~\ref{tab:app_cross_scene}). Switching from predicted to GT attributes reduces LPIPS by 0.0027 on the full image and by 0.0129 on the crop. The larger local gap shows where attribute recovery matters most in this comparison.

Figures~\ref{fig:app_insertion_1}--\ref{fig:app_insertion_3} provide visual comparisons with the source and target images, the five image editors, our result, and the Blender reference.

\section{User Study on Cross-Scene Insertion}
\label{app:user_study}

\paragraph{Task and methods.}
We evaluate 30 insertion cases with foreground objects selected from real photographs and target backgrounds drawn from both real photographs and rendered scenes.
IntrinsicLayers recovers the source object's intrinsic attributes, and IntrinsicInsert uses these predicted attributes to synthesize the object in the target background.
We compare this complete pipeline with five proprietary image-generation and editing models: Nano Banana Pro, Seedream 5.5, GPT Image 2.6 Sunburst, MiniMax H3 Image, and Kling Image 4.0.

\paragraph{Study protocol.}
Each of 66 participants evaluates all 30 cases and selects one best overall result among the six methods for each case.
Participants jointly consider background preservation, naturalness of insertion, and preservation of object attributes when making this single choice.
The study therefore yields $66\times30=1{,}980$ votes.

\begin{figure}[!htbp]
    \centering
    \includegraphics[width=\linewidth,pagebox=cropbox]{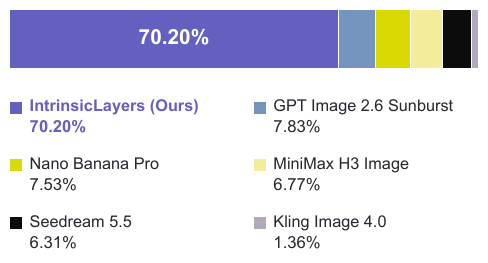}
    \caption{\textbf{User preference for cross-scene insertion from real photographs.} Vote shares over 1,980 single-choice judgments from 66 participants evaluating 30 cases. Participants jointly consider background preservation, insertion naturalness, and object-attribute preservation. The \emph{IntrinsicLayers (Ours)} label denotes the full IntrinsicLayers--IntrinsicInsert pipeline.}
    \label{fig:app_user_study}
\end{figure}

\begin{figure*}[!tp]
    \centering
    \includegraphics[width=\textwidth,pagebox=cropbox]{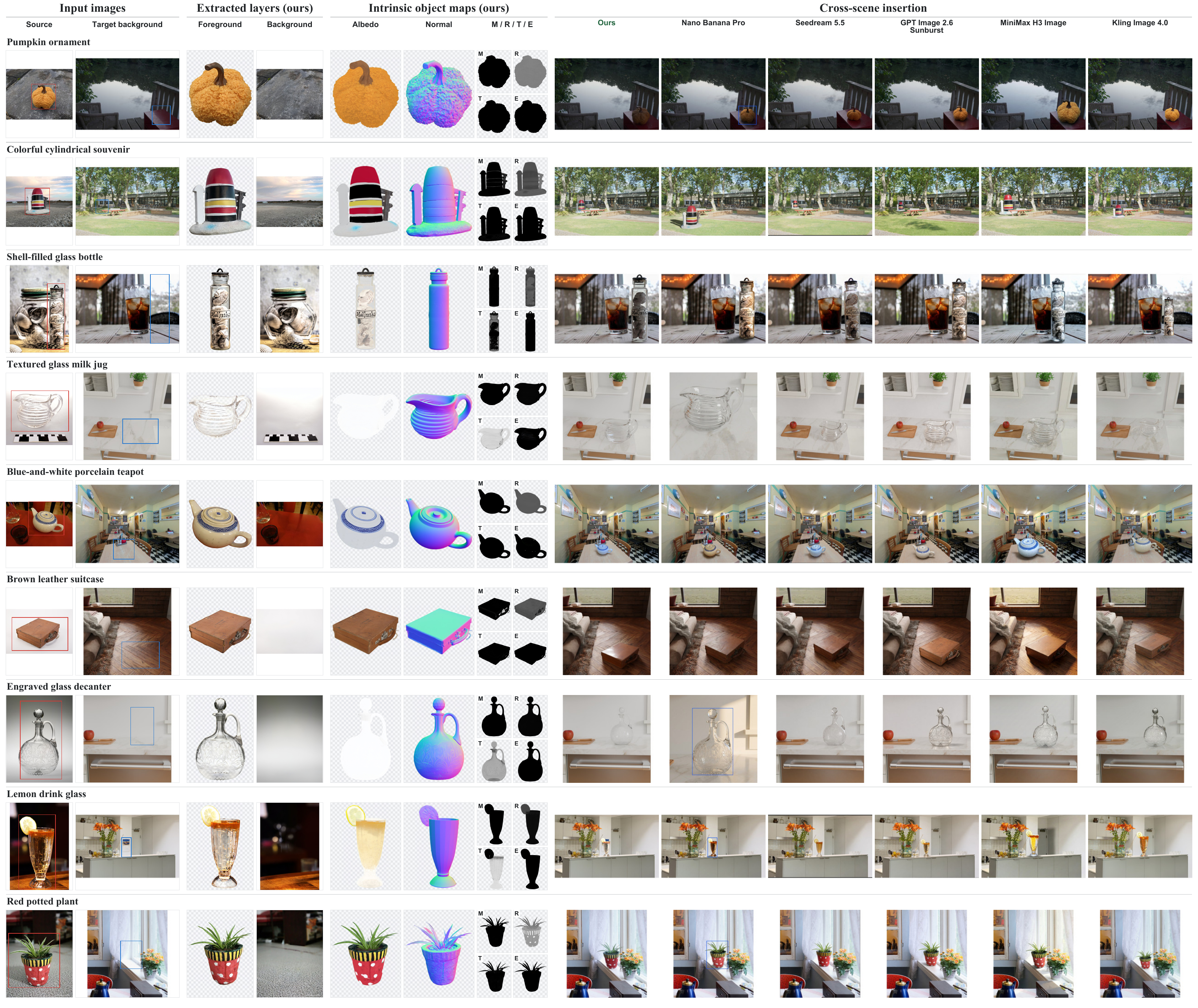}
    \caption{\textbf{Qualitative results from the cross-scene insertion user study.}}
    \label{fig:app_user_study_cases}
\end{figure*}

\paragraph{Results.}
Figure~\ref{fig:app_user_study} summarizes the vote shares.
Our pipeline receives 1,390 votes (70.20\%), compared with 155 votes (7.83\%) for GPT Image 2.6 Sunburst, the next most frequently selected method.
These results show a strong overall preference for our extraction-and-insertion pipeline on the evaluated cases.
Figure~\ref{fig:app_user_study_cases} presents nine cases from the study, showing our layer and attribute predictions alongside insertion results from all six methods.

\section{More Qualitative Results}
\label{app:more_qualitative_results}

\begin{figure*}[p]
    \centering
    \includegraphics[width=\textwidth,pagebox=cropbox]{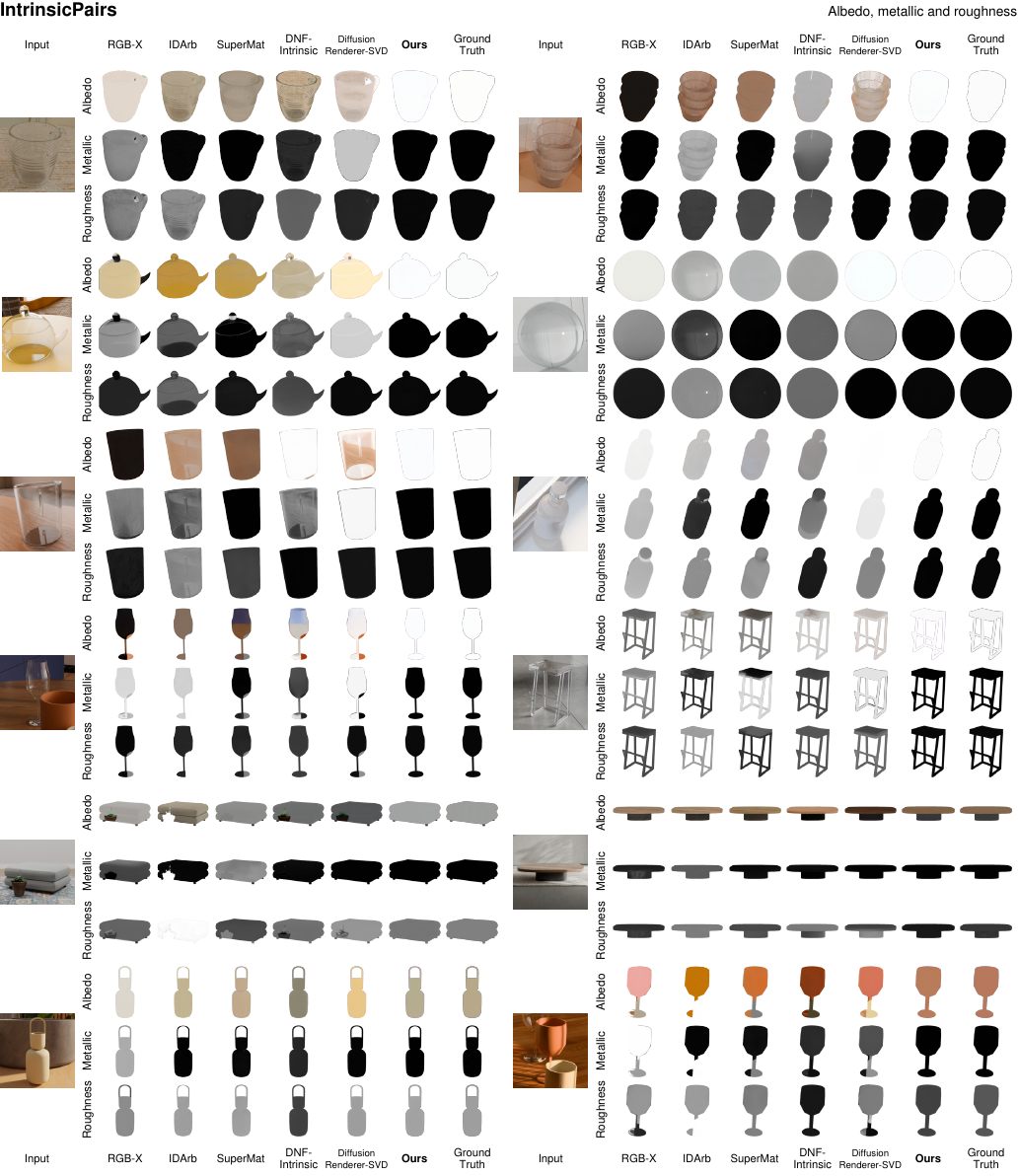}
    \caption{\textbf{More intrinsic-recovery results on IntrinsicPairs-Test.} Each example compares albedo, metalness, and roughness with the ground truth and baseline predictions. IntrinsicLayers recovers cleaner object attribute layers, reducing occluder and background appearance retained in albedo while preserving more consistent material regions. These comparisons complement Section~\ref{sec:inverse_rendering_results}.}
    \label{fig:app_amr_intrinsicpairs}
\end{figure*}

\begin{figure*}[p]
    \centering
    \includegraphics[width=\textwidth,pagebox=cropbox]{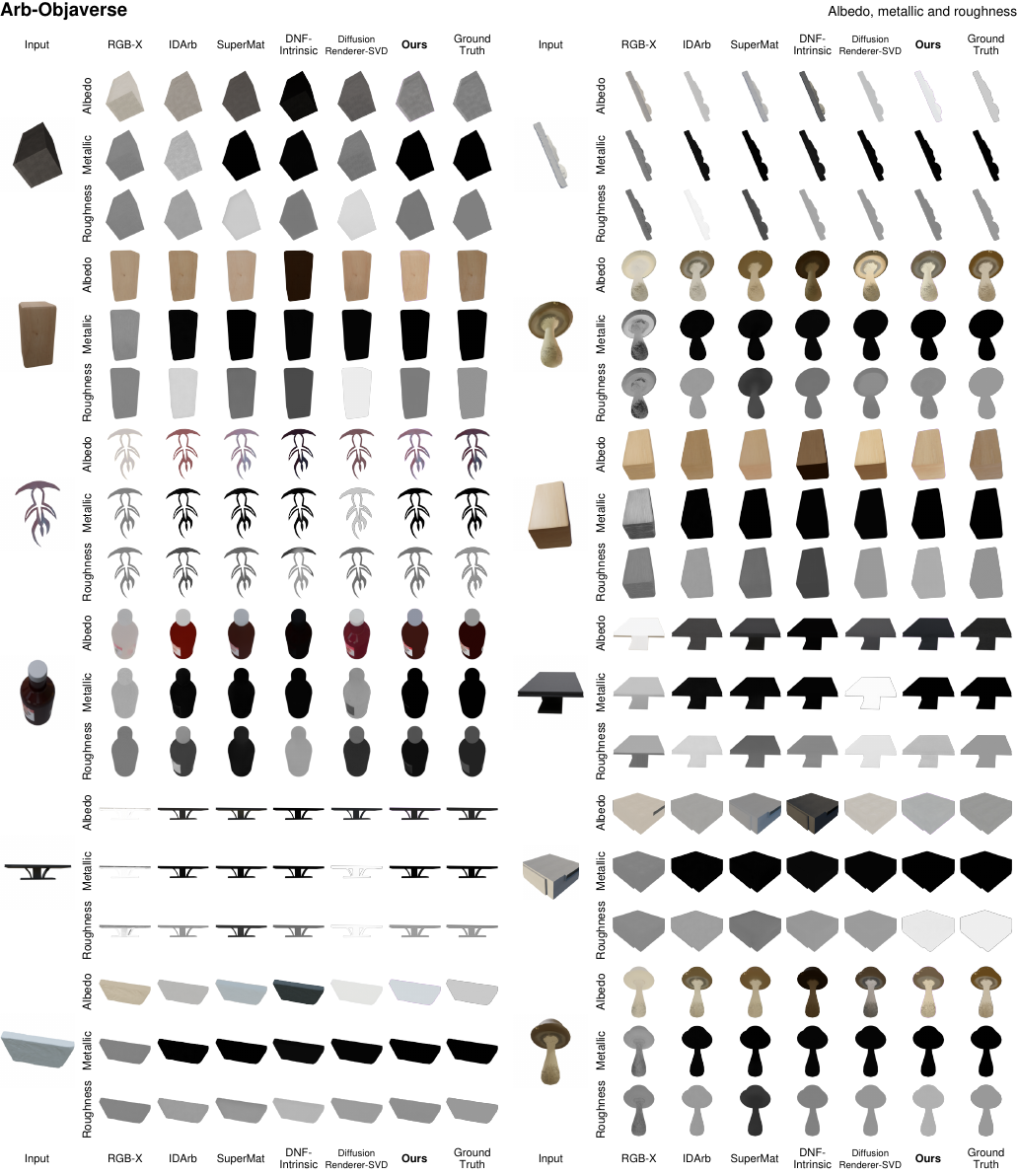}
    \caption{\textbf{More intrinsic-recovery results on Arb-Objaverse.} Additional albedo, metalness, and roughness comparisons for the public-benchmark experiment in Section~\ref{sec:public_inverse_rendering_results}. IntrinsicLayers better preserves material-region boundaries and fine surface textures, including wood grain and object patterns, yielding more detailed albedo and cleaner material maps.}
    \label{fig:app_amr_arbobjaverse}
\end{figure*}

\begin{figure*}[p]
    \centering
    \includegraphics[width=\textwidth,pagebox=cropbox]{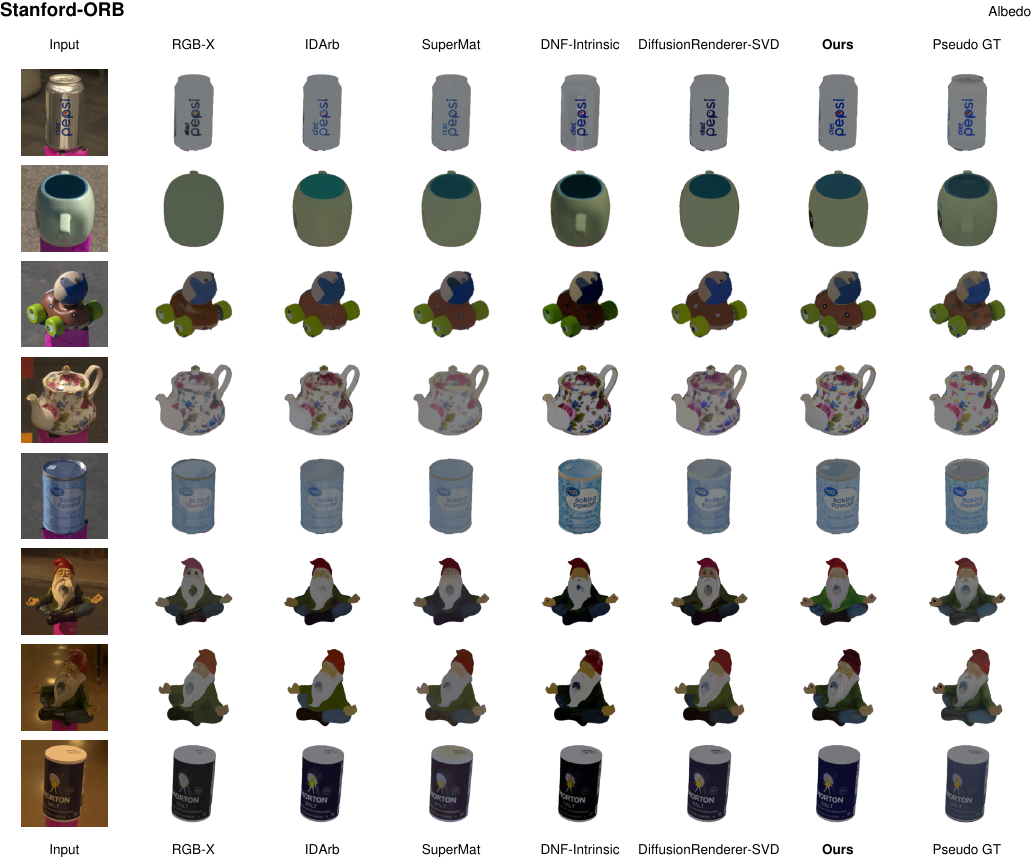}
    \caption{\textbf{More albedo results on Stanford-ORB.} Additional real-object comparisons for Section~\ref{sec:public_inverse_rendering_results}; the rightmost column shows pseudo ground truth. IntrinsicLayers preserves finer lettering, decorative patterns, and facial details while suppressing scene-dependent shading and highlights, producing clearer textures and higher-quality albedo.}
    \label{fig:app_amr_stanfordorb}
\end{figure*}

\begin{figure*}[p]
    \centering
    \includegraphics[width=\textwidth,pagebox=cropbox]{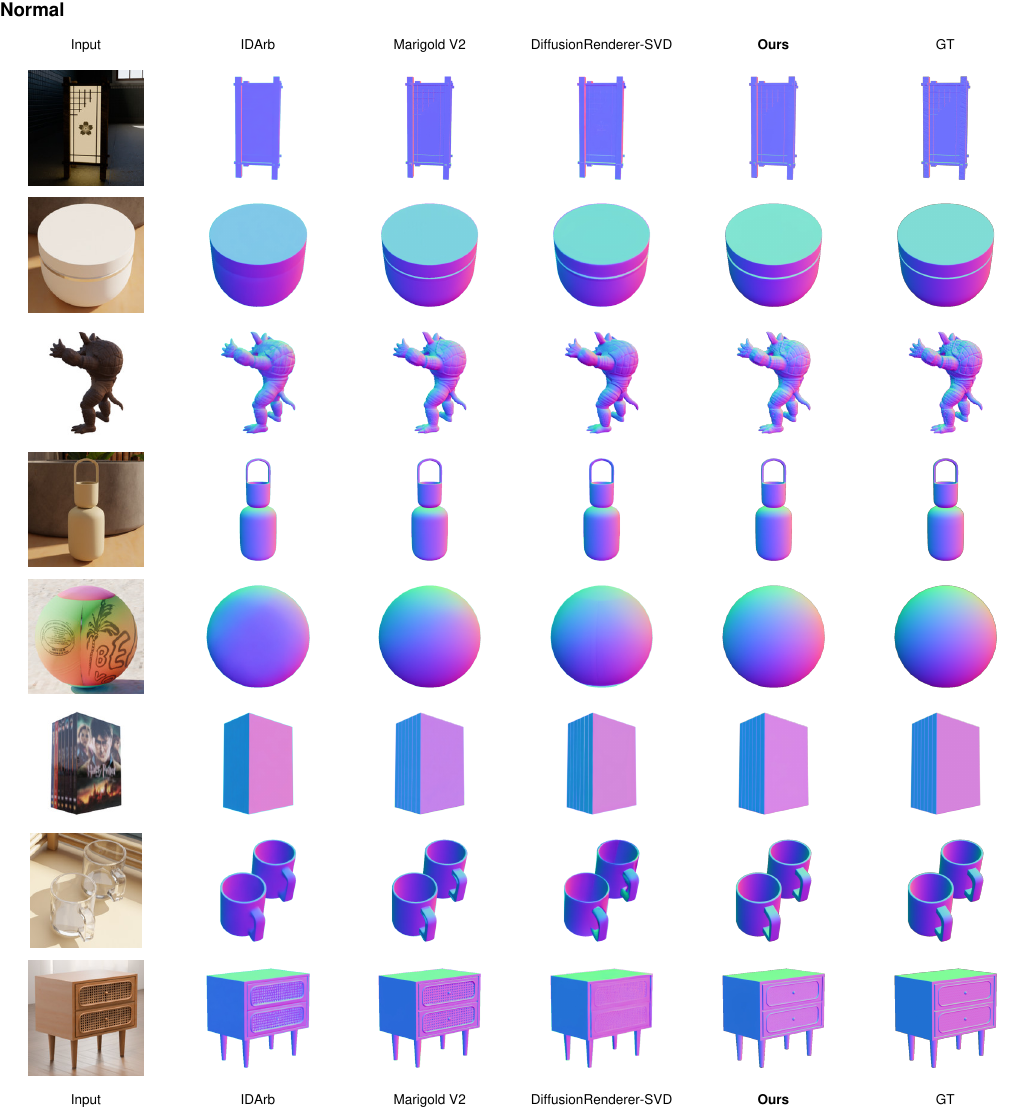}
    \caption{\textbf{More camera-space normal results.} Additional comparisons for the inverse-rendering experiment in Section~\ref{sec:qualitative_results}. IntrinsicLayers preserves clearer object boundaries and fine geometric structure, including the book spines, while reducing appearance-induced artifacts on the cabinet surfaces.}
    \label{fig:app_normal_1}
\end{figure*}

\begin{figure*}[p]
    \centering
    \includegraphics[width=\textwidth,pagebox=cropbox]{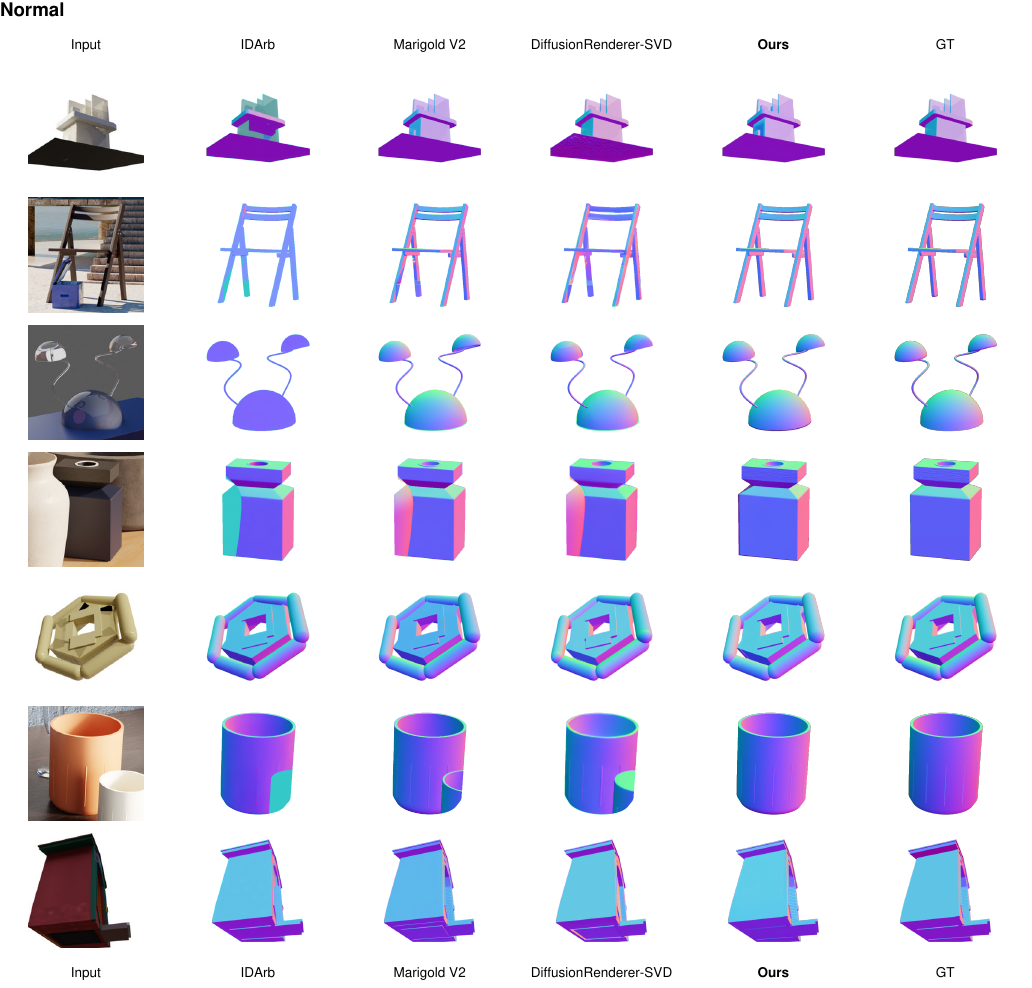}
    \caption{\textbf{More normal results with partially occluded objects.} Further comparisons extending Figure~\ref{fig:normal_comparison}. IntrinsicLayers recovers more complete target geometry and clearer planar boundaries, including the continuous cylindrical surface behind the foreground cup, with normals that more closely follow the ground-truth surface orientation.}
    \label{fig:app_normal_2}
\end{figure*}

Figures~\ref{fig:app_amr_intrinsicpairs} and~\ref{fig:app_amr_arbobjaverse} show additional albedo, metalness, and roughness comparisons on IntrinsicPairs-Test and Arb-Objaverse, respectively, for Section~\ref{sec:quantitative_results}.
Figure~\ref{fig:app_amr_stanfordorb} shows albedo on Stanford-ORB with pseudo ground truth, and Figures~\ref{fig:app_normal_1} and~\ref{fig:app_normal_2} extend the camera-space normal comparisons in Figure~\ref{fig:normal_comparison}.

For occluded and transparent targets, IntrinsicLayers recovers intrinsic layers with fewer residuals from the occluder or background while preserving material boundaries, surface textures, and fine albedo details.

\begin{figure*}[p]
    \centering
    \includegraphics[width=\textwidth,pagebox=cropbox]{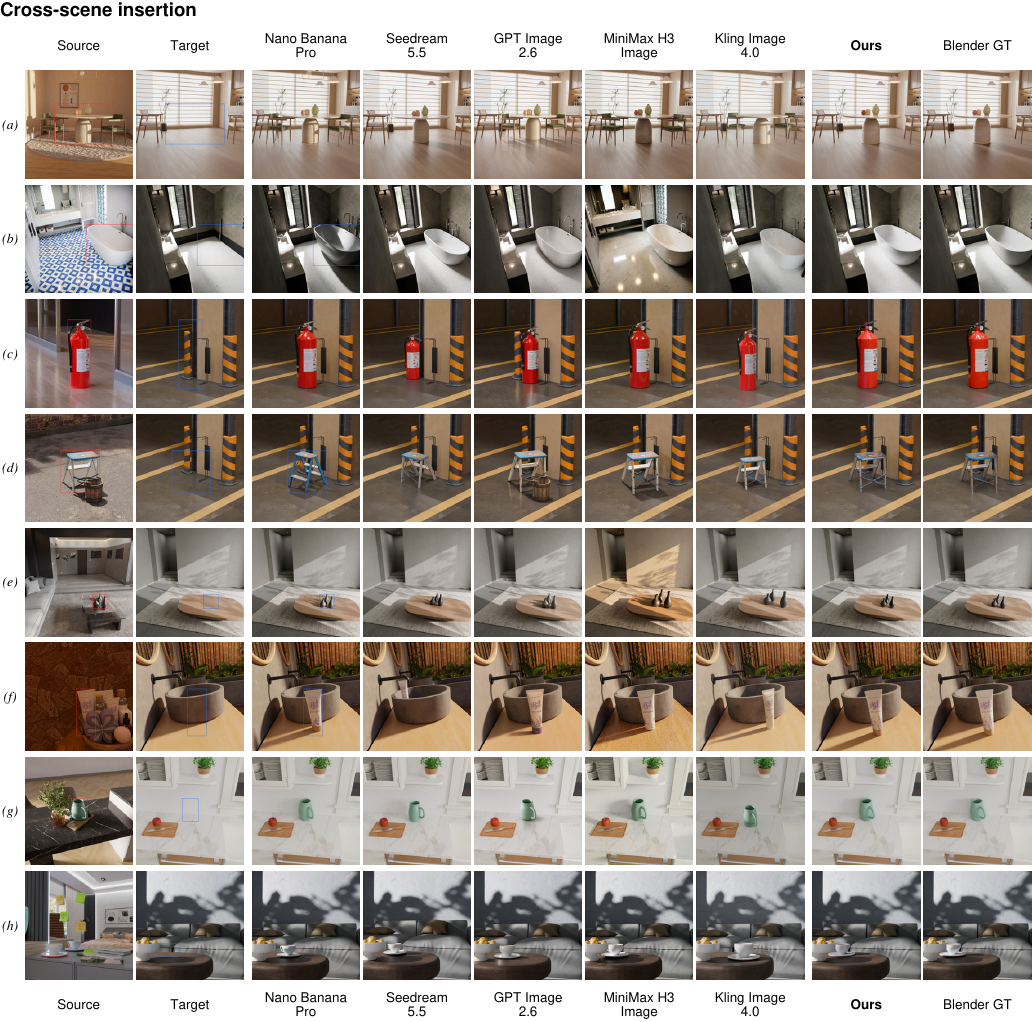}
    \caption{\textbf{Cross-scene insertion comparisons (a--h).} Qualitative results for the rendered-pair experiment in Section~\ref{app:cross_scene_rendered}. Our pipeline preserves object identity and placement while adapting appearance to the target scene. The stool example (d) illustrates transfer of the selected object without carrying over its foreground occluder.}
    \label{fig:app_insertion_1}
\end{figure*}

\begin{figure*}[p]
    \centering
    \includegraphics[width=\textwidth,pagebox=cropbox]{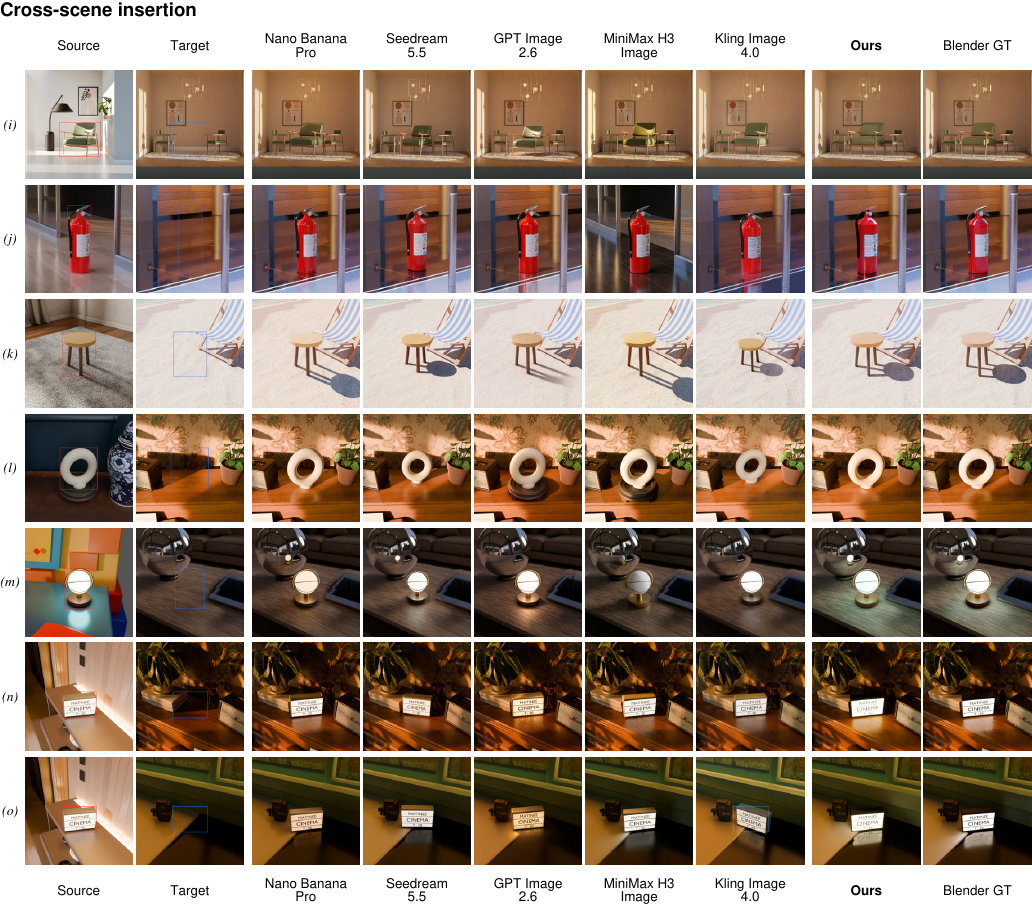}
    \caption{\textbf{Cross-scene insertion comparisons (i--o).} Additional examples from the experiment in Section~\ref{app:cross_scene_rendered}, illustrating adaptation to target lighting, cast shadows (k), and illumination and reflections produced by emissive objects (m--o). Blender renderings provide references for these object--environment interactions.}
    \label{fig:app_insertion_2}
\end{figure*}

\begin{figure*}[p]
    \centering
    \includegraphics[width=\textwidth,pagebox=cropbox]{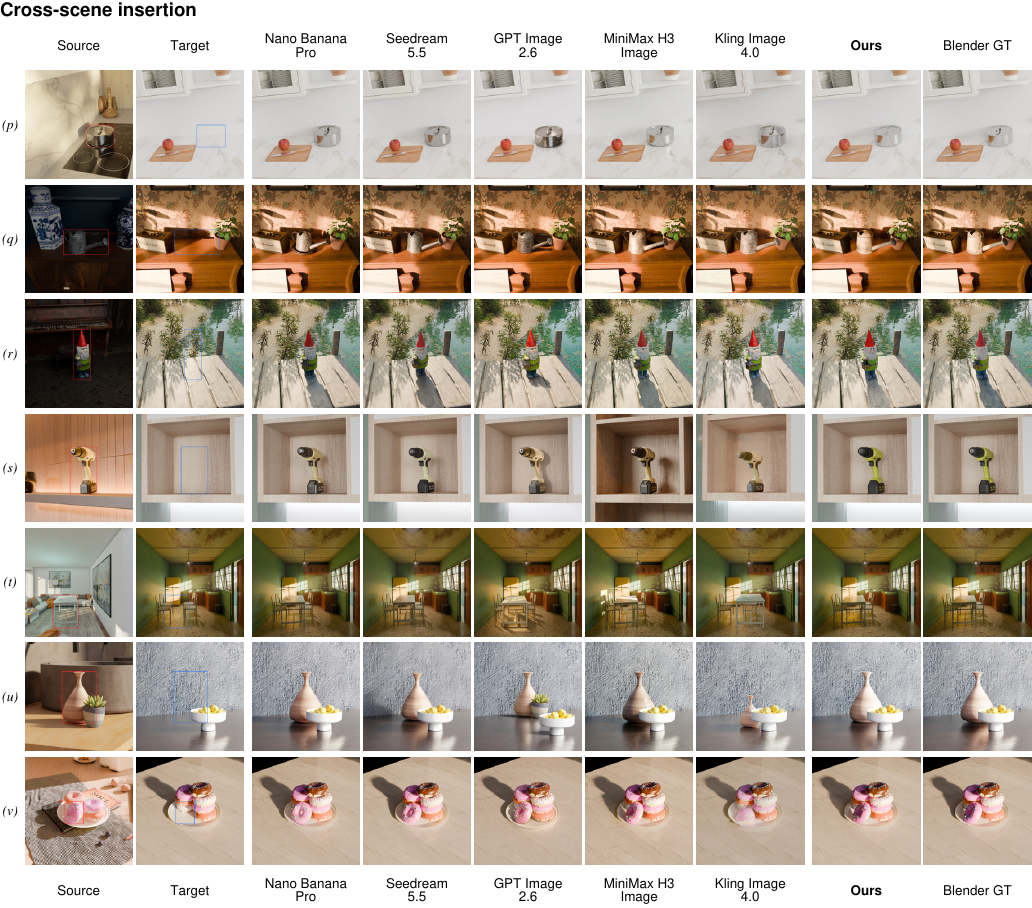}
    \caption{\textbf{Cross-scene insertion comparisons (p--v).} Further results for Section~\ref{app:cross_scene_rendered}. Our pipeline adapts reflective appearance and object shading while preserving the target-scene structure. In (u), the vase is transferred without the source-scene plant and is correctly placed behind the target-scene bowl.}
    \label{fig:app_insertion_3}
\end{figure*}

\paragraph{Cross-scene insertion.}
Figures~\ref{fig:app_insertion_1}--\ref{fig:app_insertion_3} complement the rendered-pair evaluation in Section~\ref{app:cross_scene_rendered} and Table~\ref{tab:app_cross_scene}, extending the reinsertion examples in Figure~\ref{fig:qualitative_cases}.
The red box selects the source object, and the blue box specifies its placement in the target background; each row compares five image editors and our pipeline against a Blender reference.
The examples show preservation of object identity and placement while adapting shading, shadows, and emissive effects to the target scene, including transfer of partially occluded source objects without their occluders.

\begin{figure*}[p]
    \centering
    \includegraphics[width=\textwidth,pagebox=cropbox]{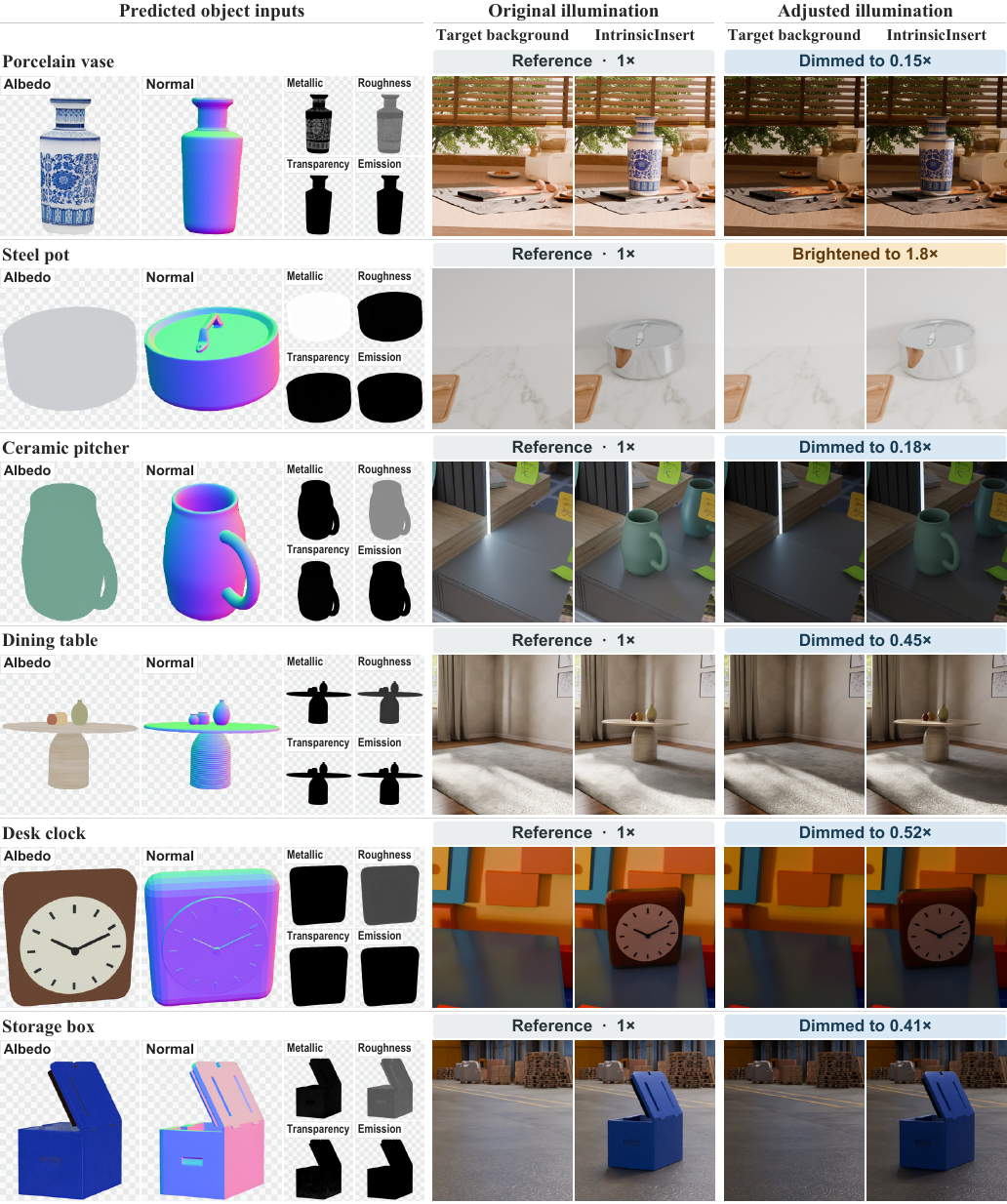}
    \caption{\textbf{Response to target-background illumination.} Each row pairs the same predicted intrinsic inputs with reference and adjusted target backgrounds and shows the corresponding IntrinsicInsert outputs. The inserted objects adapt to darker or brighter surroundings while preserving their structure and texture; the labeled multipliers are relative to the reference setting ($1\times$).}
    \label{fig:app_lighting_sensitivity}
\end{figure*}

\paragraph{Response to target-background illumination.}
Figure~\ref{fig:app_lighting_sensitivity} reuses predicted object attributes under original and adjusted background illumination.
IntrinsicInsert adapts object brightness and shading to the background RGB while preserving structure and texture.

\begin{figure*}[p]
    \centering
    \includegraphics[width=\textwidth,pagebox=cropbox]{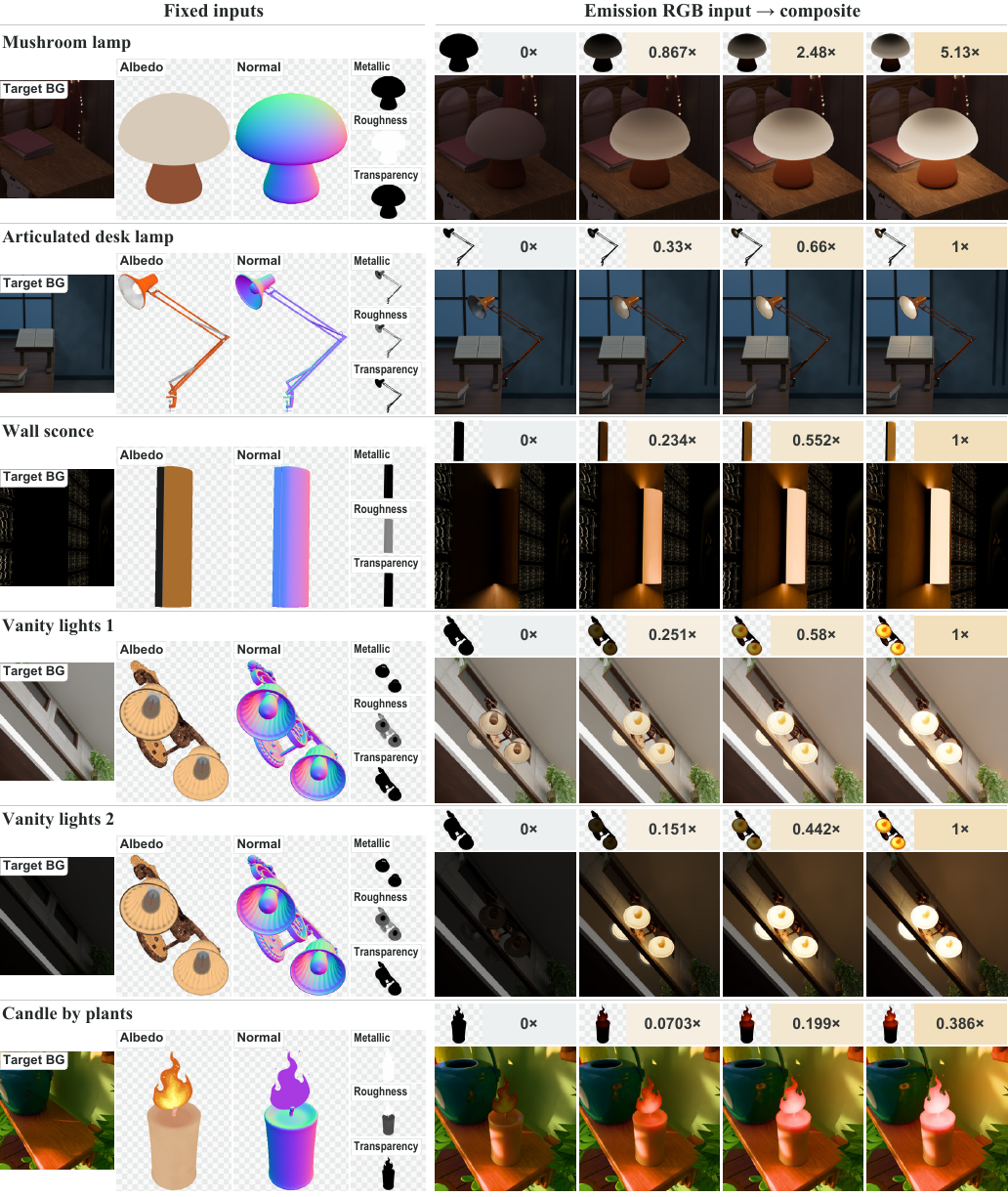}
    \caption{\textbf{Attribute control in IntrinsicInsert.} Within each row, the target background, albedo, normals, metalness, roughness, and transmission are fixed, while the emission RGB input is varied at the labeled strengths. Each edited emission reference is shown above its composite, demonstrating control over the object's luminous appearance and its illumination of nearby surfaces.}
    \label{fig:app_emission_control}
\end{figure*}

\paragraph{Attribute-controlled insertion.}
Figure~\ref{fig:app_emission_control} demonstrates control of IntrinsicInsert through its emission attribute.
With the target background and other attributes fixed, editing emission changes the object's luminous appearance and its illumination of nearby surfaces.

\begin{figure*}[p]
    \centering
    \includegraphics[width=\textwidth,pagebox=cropbox]{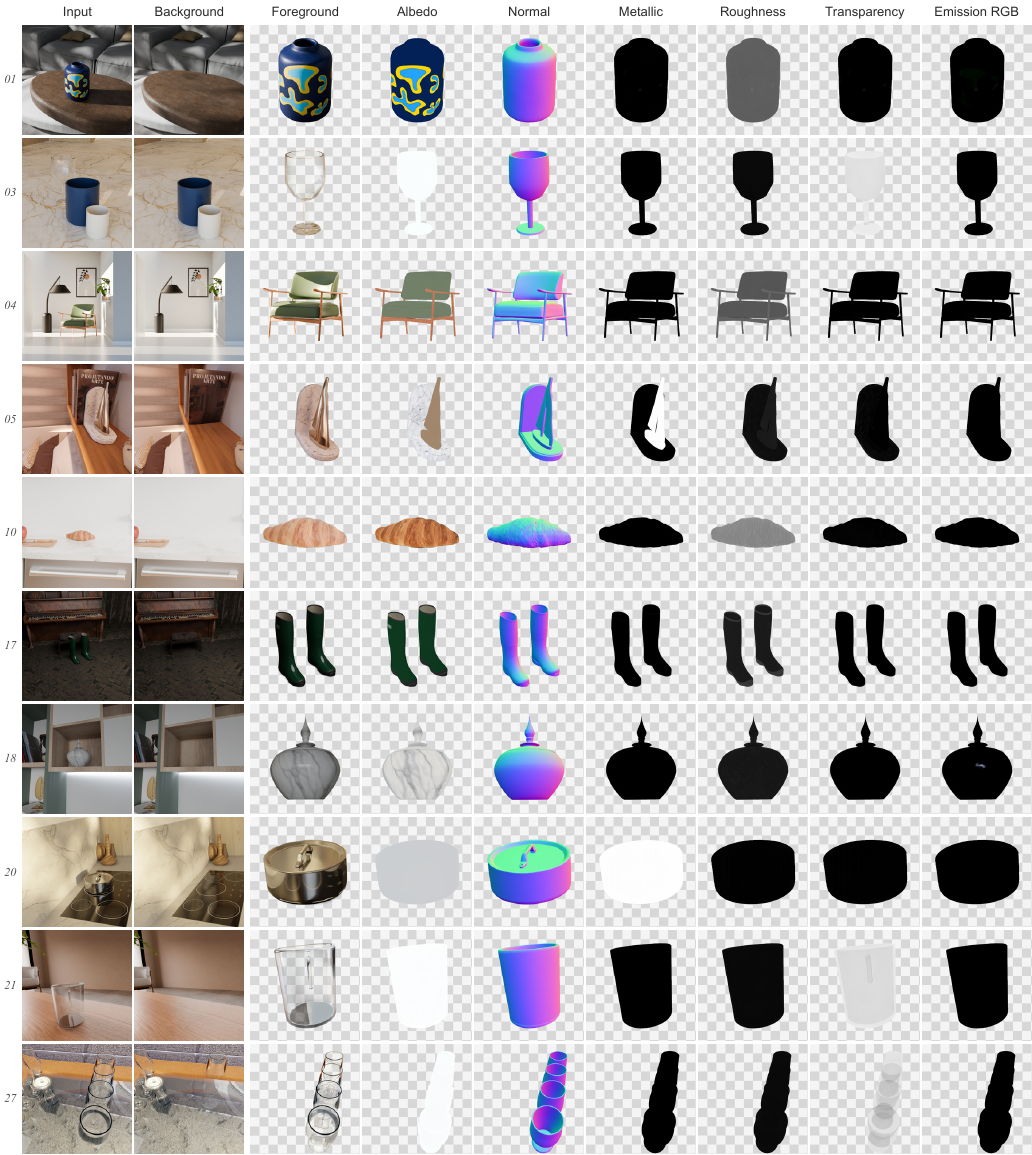}
    \caption{\textbf{Joint layer and intrinsic decomposition by IntrinsicLayers.} Each example shows the recovered background, foreground, and aligned albedo, camera-space normals, metalness, roughness, transmission, and RGB emission. The partially occluded wine glass (03) illustrates completion of the target projection and its intrinsic maps.}
    \label{fig:app_decomposition_1}
\end{figure*}

\begin{figure*}[p]
    \centering
    \includegraphics[width=\textwidth,pagebox=cropbox]{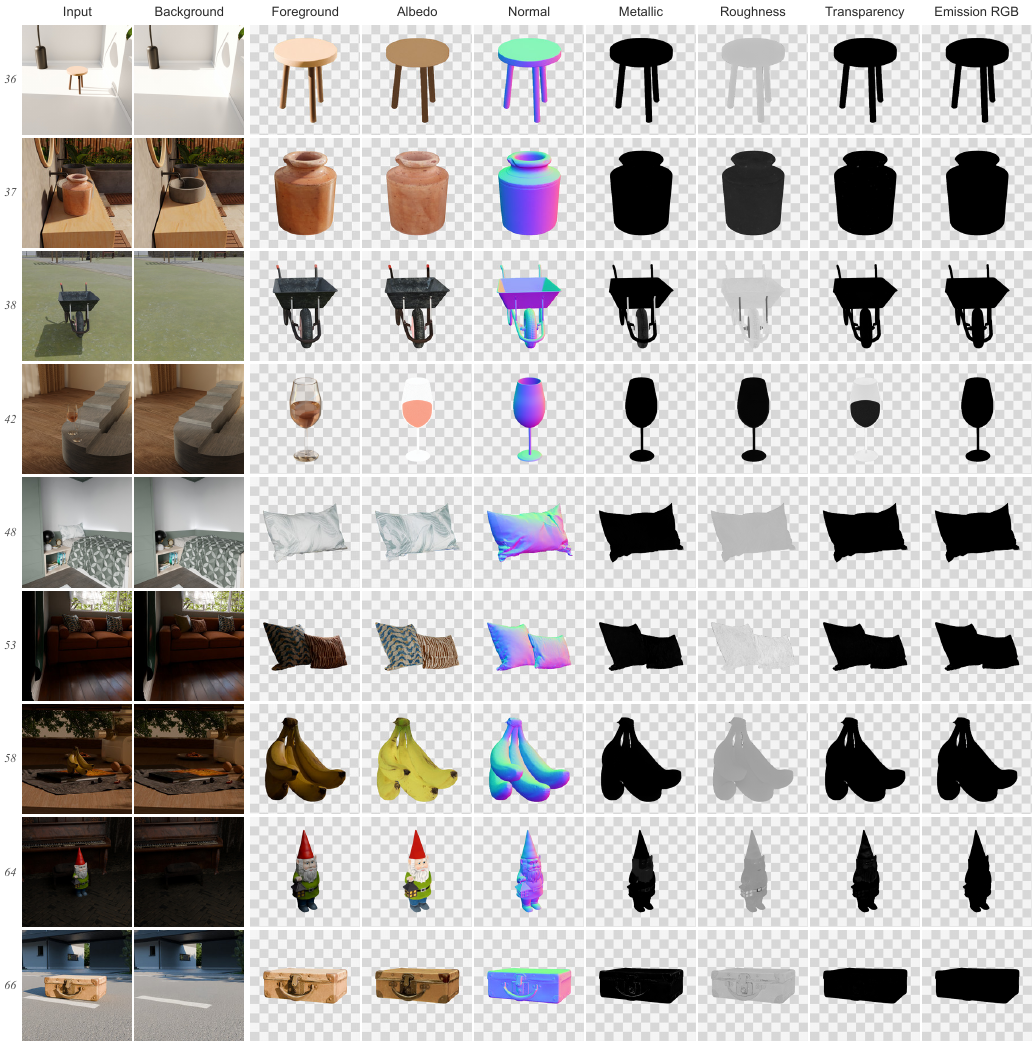}
    \caption{\textbf{Joint decomposition of textured and multi-part objects.} Albedo preserves fabric patterns and surface details, while normal maps capture folds, relief, and object structure. The recovered backgrounds show the scenes after target removal, alongside the extracted foregrounds and their material attributes.}
    \label{fig:app_decomposition_2}
\end{figure*}

\begin{figure*}[p]
    \centering
    \includegraphics[width=\textwidth,pagebox=cropbox]{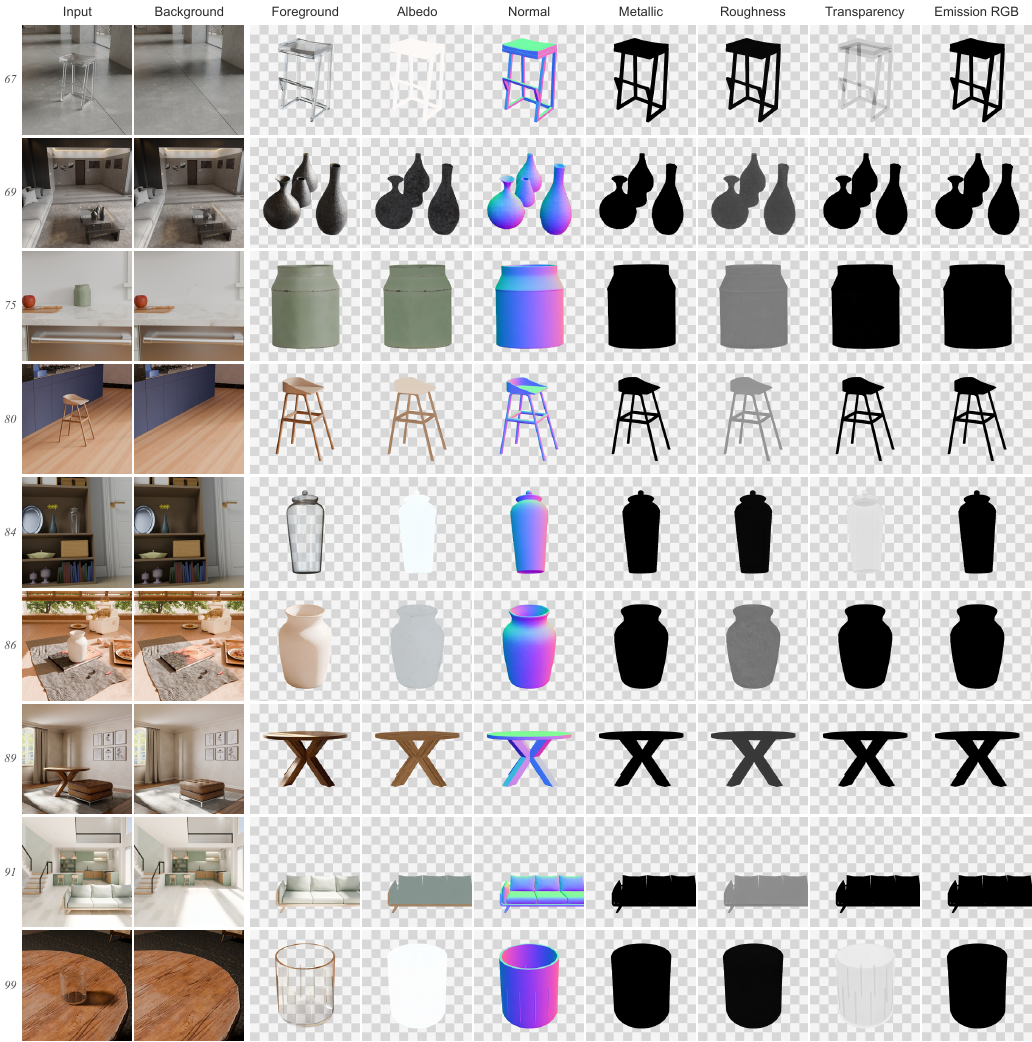}
    \caption{\textbf{Further joint decomposition results, including transmissive furniture and vessels.} Separate foreground and transmission outputs accompany aligned geometry and material maps. The partially occluded table (89) illustrates recovery of the complete target projection, including the legs hidden in the input image.}
    \label{fig:app_decomposition_3}
\end{figure*}

\begin{figure*}[p]
    \centering
    \includegraphics[width=\textwidth,pagebox=cropbox]{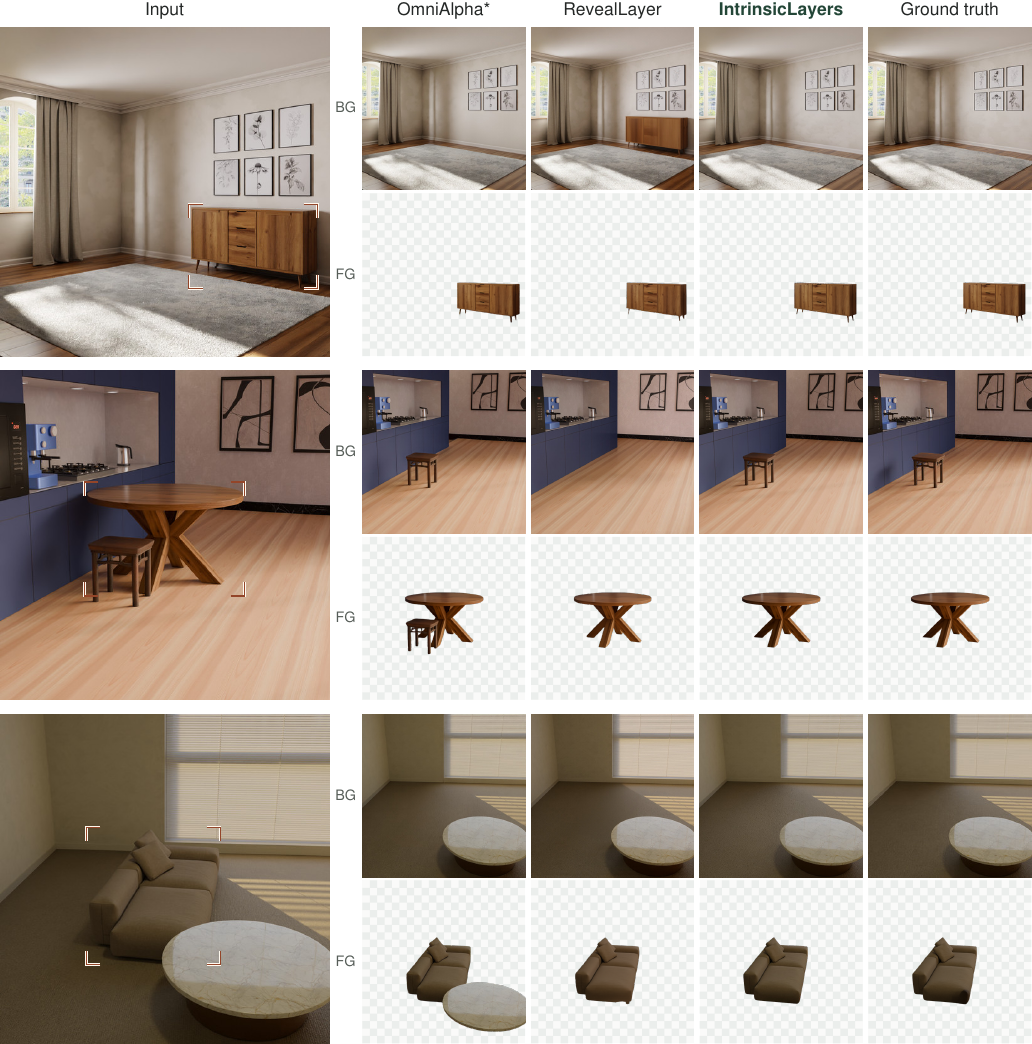}
    \caption{\textbf{Layer decomposition in indoor scenes.} Each example compares the recovered background (BG) and foreground (FG) with the rendered ground truth. IntrinsicLayers isolates the selected furniture while preserving surrounding objects, such as the stool beside the dining table. OmniAlpha$^*$ uses the configuration in Table~\ref{tab:layer_results}.}
    \label{fig:app_layer_comparison_1}
\end{figure*}

\begin{figure*}[p]
    \centering
    \includegraphics[width=\textwidth,pagebox=cropbox]{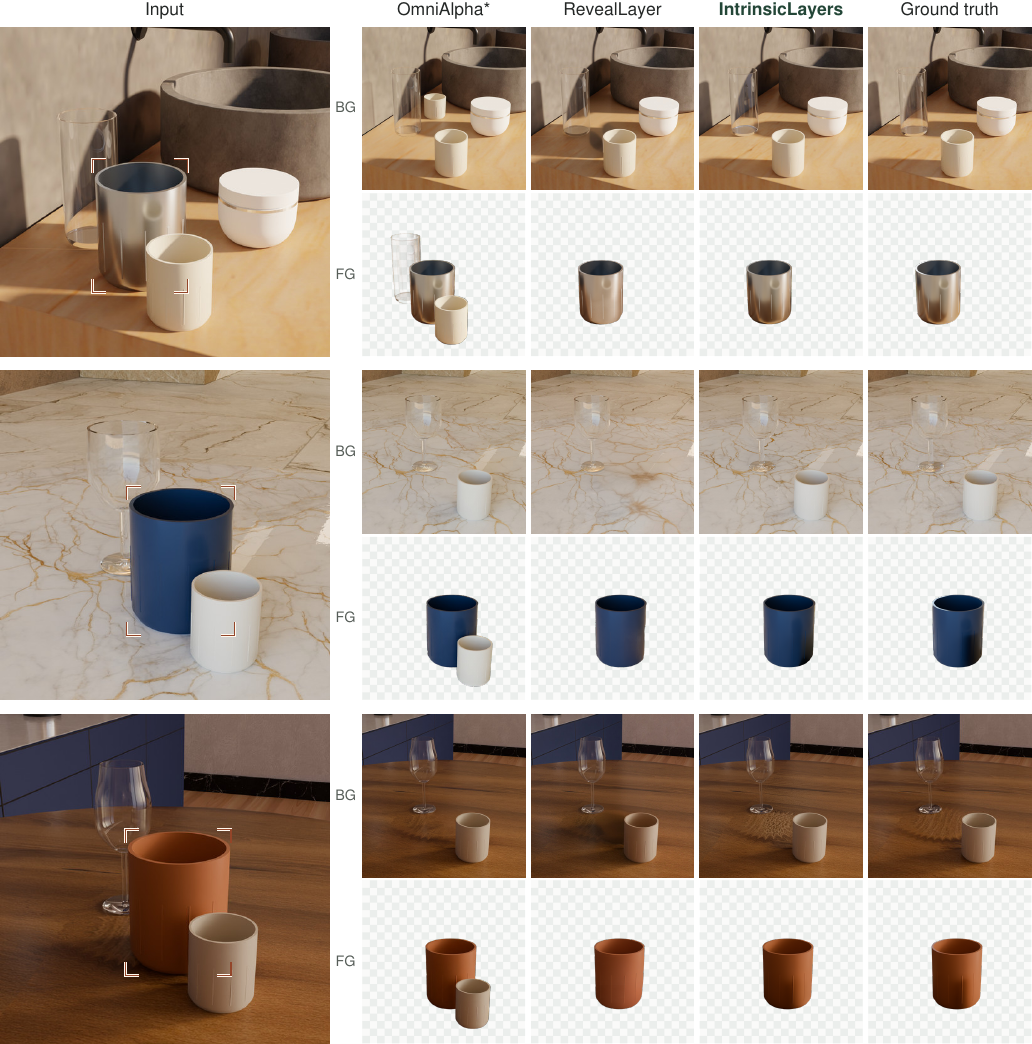}
    \caption{\textbf{Layer decomposition of partially occluded cups.} IntrinsicLayers isolates each target cup from the surrounding vessels, completes its foreground behind the occluder, and retains the neighboring vessels in the background.}
    \label{fig:app_layer_comparison_2}
\end{figure*}

\begin{figure*}[p]
    \centering
    \includegraphics[width=\textwidth,pagebox=cropbox]{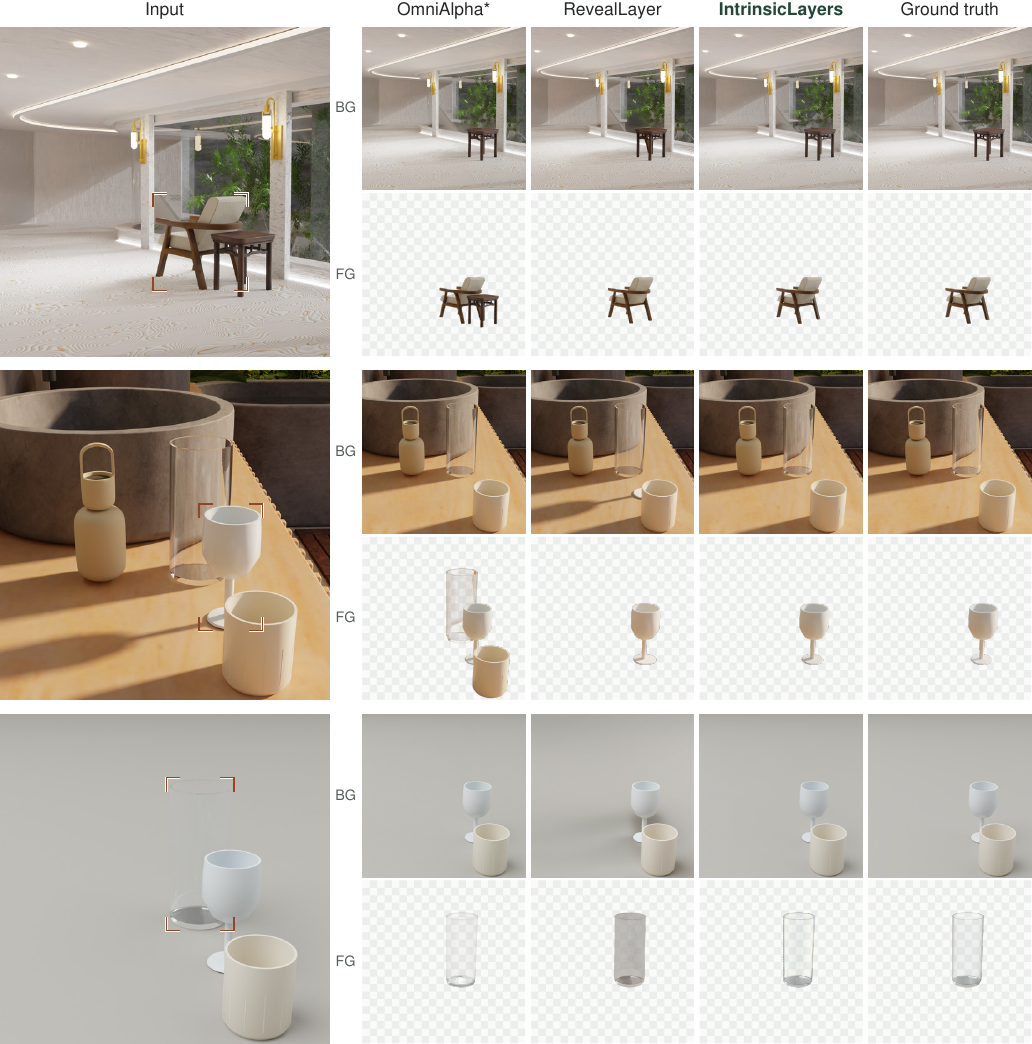}
    \caption{\textbf{Layer decomposition of furniture and glassware.} The top example separates the selected chair from the adjacent table. The glass example at the bottom illustrates foreground extraction for a transparent target in a low-contrast scene.}
    \label{fig:app_layer_comparison_3}
\end{figure*}

\begin{figure*}[p]
    \centering
    \includegraphics[width=\textwidth,pagebox=cropbox]{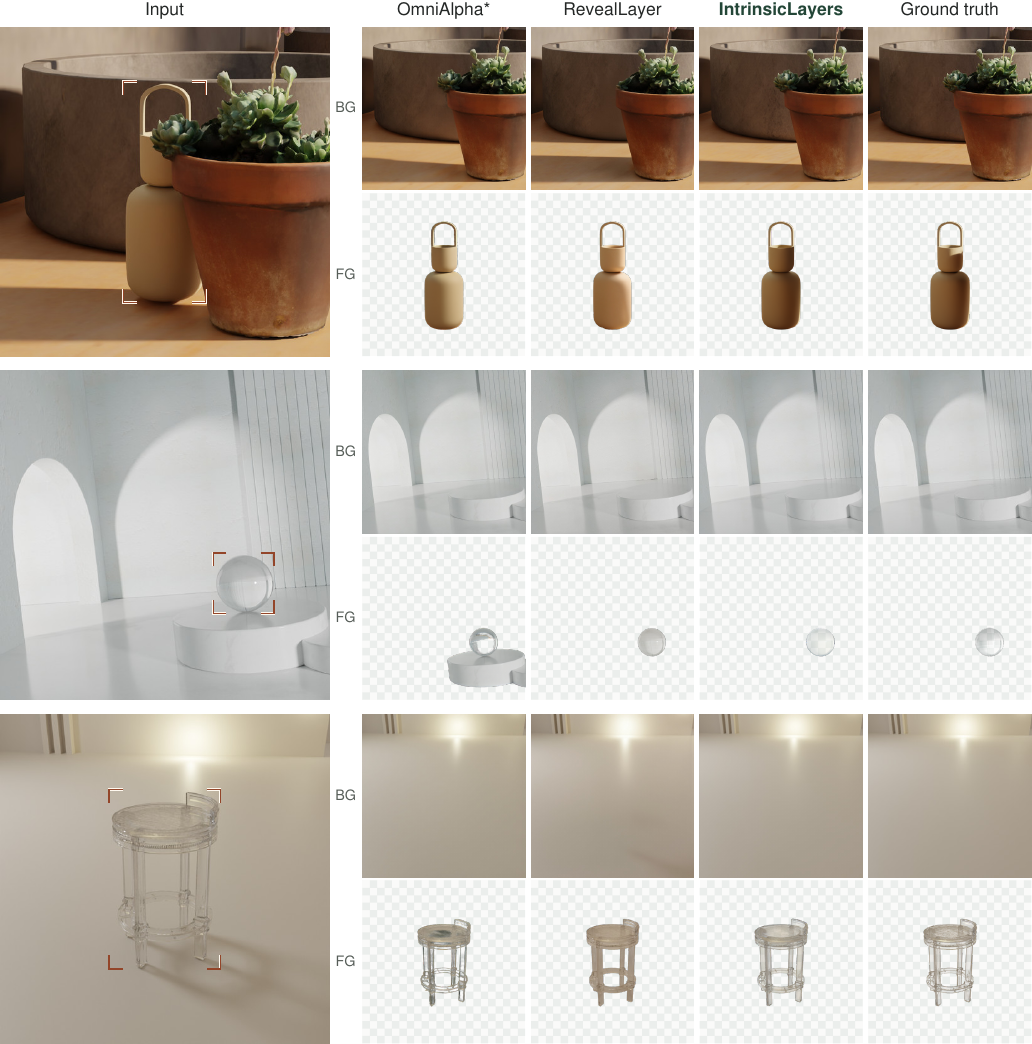}
    \caption{\textbf{Further layer decomposition with occlusion and transparency.} The top example shows completion of the selected object behind a plant pot. The lower examples show foregrounds and target-absent backgrounds for a glass sphere and a transparent stool.}
    \label{fig:app_layer_comparison_4}
\end{figure*}

\begin{figure*}[p]
    \centering
    \includegraphics[width=\textwidth,pagebox=cropbox]{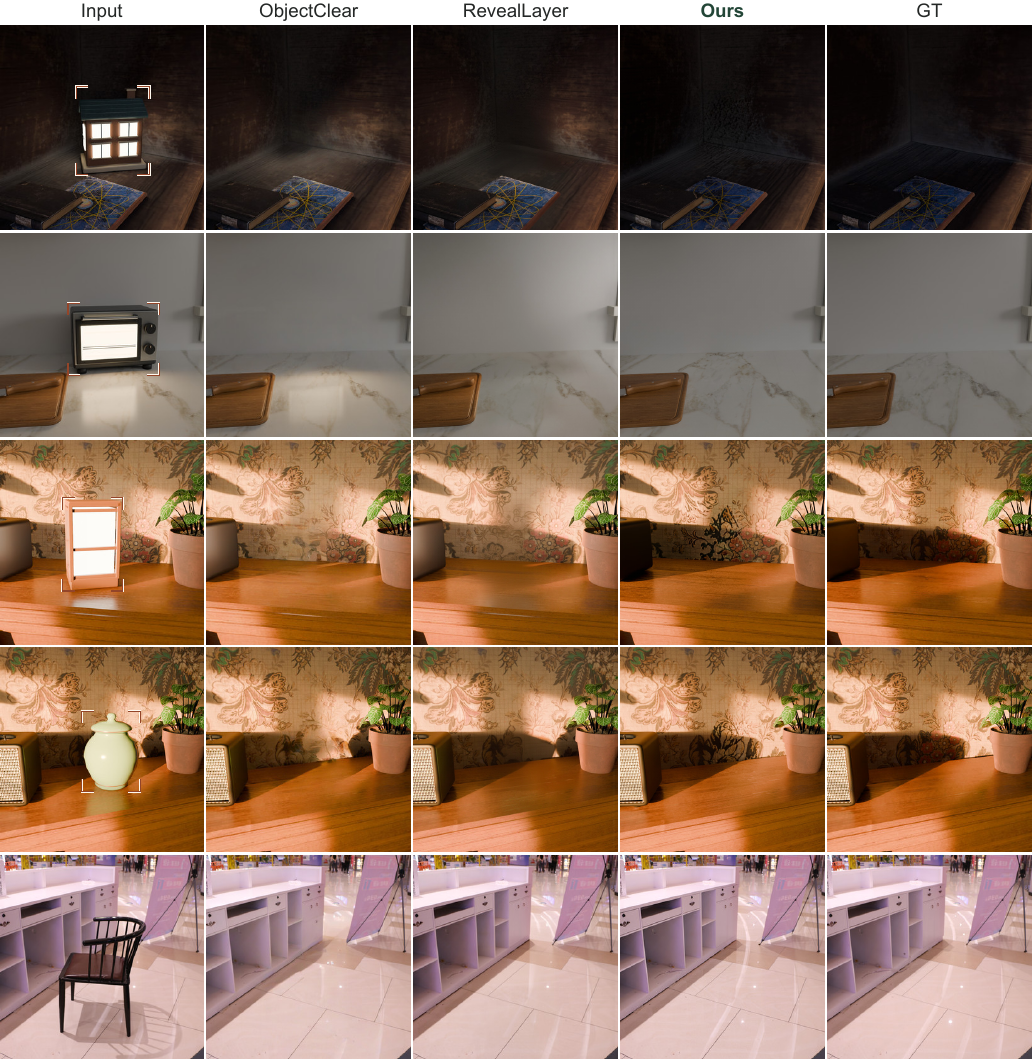}
    \caption{\textbf{Object removal and scene effects.} Each row compares ObjectClear, RevealLayer, and IntrinsicLayers with the ground-truth background. The lamp and oven examples illustrate changes in illumination beyond the removed object, and the chair example shows the recovered floor.}
    \label{fig:app_removal_comparison}
\end{figure*}

\begin{figure*}[p]
    \centering
    \includegraphics[width=\textwidth,pagebox=cropbox]{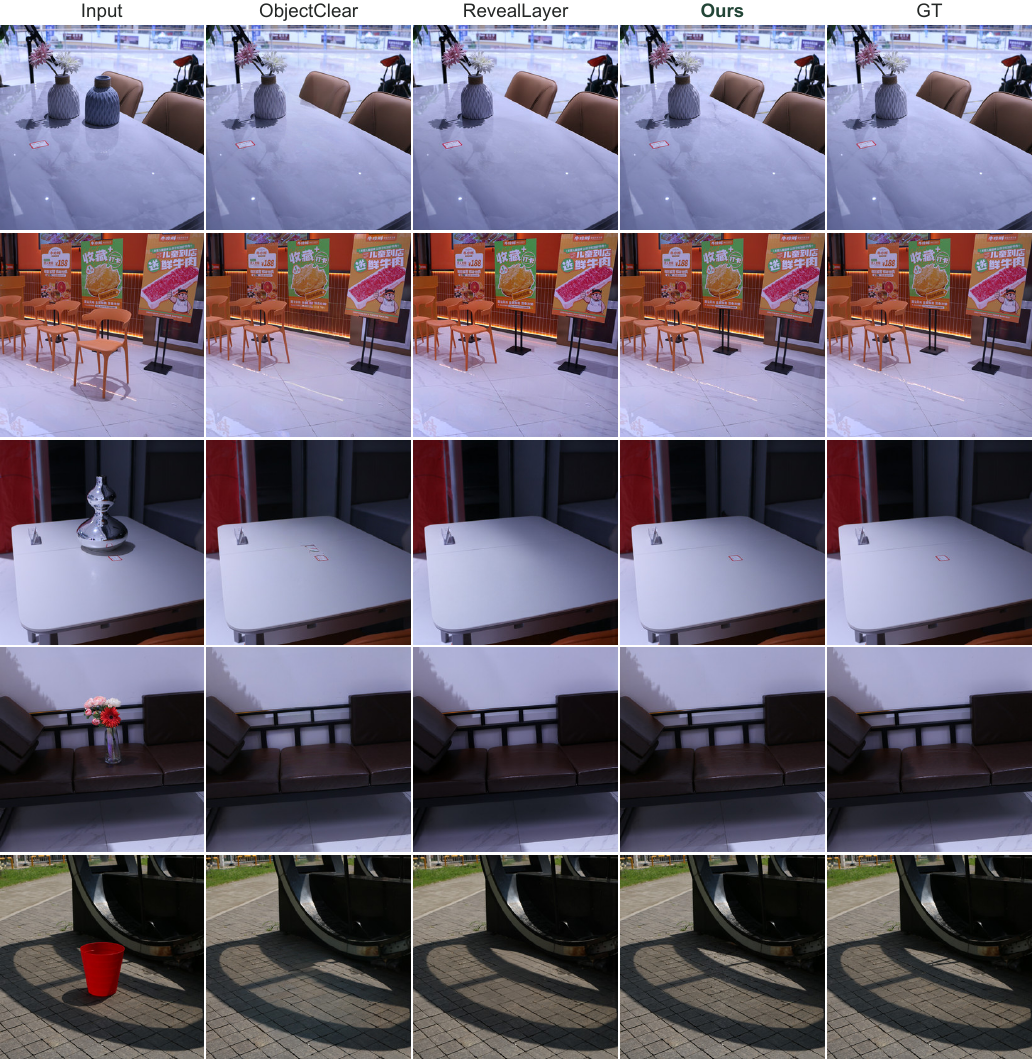}
    \caption{\textbf{Further object removal comparisons.} The examples show removal in indoor and outdoor scenes, with ground-truth backgrounds in the rightmost column.}
    \label{fig:app_removal_comparison_2}
\end{figure*}

\paragraph{Joint layer and intrinsic decomposition.}
Figures~\ref{fig:app_decomposition_1}--\ref{fig:app_decomposition_3} show joint background, foreground, and intrinsic predictions from IntrinsicLayers, complementing the layer and attribute experiments in Section~\ref{sec:qualitative_results}.
The outputs follow the representation in Section~\ref{sec:joint_decomposition}; the column labeled \emph{Transparency} visualizes the predicted transmission $T$.
Figures~\ref{fig:app_layer_comparison_1}--\ref{fig:app_layer_comparison_4} extend the layer decomposition results in Table~\ref{tab:layer_results}, and Figures~\ref{fig:app_removal_comparison} and~\ref{fig:app_removal_comparison_2} extend the object removal results in Table~\ref{tab:removal_results}.

\clearpage
\renewcommand{\refname}{Supplementary References}
\end{document}